\pdfoutput=1

\documentclass[11pt]{article}

\usepackage{emnlp2021}

\usepackage{times}
\usepackage{latexsym}

\usepackage[T1]{fontenc}

\usepackage[utf8]{inputenc}

\usepackage{microtype}

\usepackage{url}
\usepackage{makecell}
\usepackage{booktabs} 
\usepackage{arydshln} 
\usepackage{multirow}
\usepackage{graphicx}
\usepackage{subcaption}
\usepackage{dashbox}
\usepackage{xcolor}
\usepackage{xspace}
\usepackage{enumerate}

\newcommand{\tablefont}{\small}

\newcommand{\tdrawer}{\mbox{\sc{TDrawer}}\xspace}
\newcommand{\tguesser}{\mbox{\sc{TGuesser}}\xspace}
\newcommand{\ind}{\mbox{\sc{Ind}}\xspace}
\newcommand{\ood}{\mbox{\sc{Ood}}\xspace}
\newcommand{\oov}{\mbox{\sc{OoV}}\xspace}
\newcommand{\boldheader}[1]{\vspace{0.15cm}\noindent\textbf{#1}:}

\newcommand{\imsitu}{\mbox{imSitu}\xspace}

\title{Iconary: A Pictionary-Based Game for Testing Multimodal Communication with Drawings and Text}

\newcommand{\aspace}{\hspace{0.5cm}\hfill}
\newcommand{\ai}{$^*$}
\newcommand{\upen}{$^{\dagger}$}
\newcommand{\gist}{$^\Gamma$}
\newcommand{\uw}{$^+$}

\author{Christopher Clark\ai \aspace Jordi Salvador\ai  \aspace Dustin Schwenk\ai  \aspace Derrick Bonafilia\ai \\ \bf{Mark Yatskar\upen \aspace Eric Kolve\ai \aspace Alvaro Herrasti\ai \aspace Jonghyun Choi\ai\gist} \\ \bf{Sachin Mehta\uw \aspace Sam Skjonsberg\ai \aspace Carissa Schoenick\ai \aspace Aaron Sarnat\ai} \\ \textbf{Hannaneh Hajishirzi\ai\uw \aspace Aniruddha Kembhavi\ai\uw \aspace Oren Etzioni\ai \aspace Ali Farhadi\ai\uw}
\\
\hspace*{0.2cm}{\ai}Allen Institute for AI \aspace {\uw}University of Washington\hspace*{0.6cm}\\
\hspace*{0.2cm}{\upen}University of Pennsylvania \aspace {\gist}Gwangju Institute of Science and Technology\hspace*{0.6cm}\\

\href{mailto:chrisc@allenai.org}{\tt chrisc@allenai.org}
}

\begin{document}
\maketitle
\begin{abstract}
\renewcommand\UrlFont{\color{red}\rmfamily\itshape}
Communicating with humans is challenging for AIs because it requires a shared understanding of the world, complex semantics (e.g., metaphors or analogies), and at times multi-modal gestures (e.g., pointing with a  finger, or an arrow in a diagram).
We investigate these challenges in the context of Iconary, a collaborative game of drawing and guessing based on Pictionary, that poses a novel challenge for the research community.
In Iconary, a Guesser tries to identify a phrase that a Drawer is drawing by composing icons, and the Drawer iteratively revises the drawing to help the Guesser in response.
This back-and-forth often uses canonical scenes, visual metaphor, or icon compositions to express challenging words, making it an ideal test for mixing language and visual/symbolic communication in AI.
We propose models to play Iconary and train them on over 55,000 games between human players. 
Our models are skillful players and are able to employ world knowledge in language models to play with words unseen during training.
Elite human players outperform our models, particularly at the drawing task, leaving an important gap for future research to address.
We release our dataset, code, and evaluation setup as a challenge to the community at \href{http://www.github.com/allenai/iconary}{\tt github.com/allenai/iconary}.

\end{abstract}

\section{Introduction}

\begin{figure*}[!t]
    \centering
    \includegraphics[width=\textwidth]{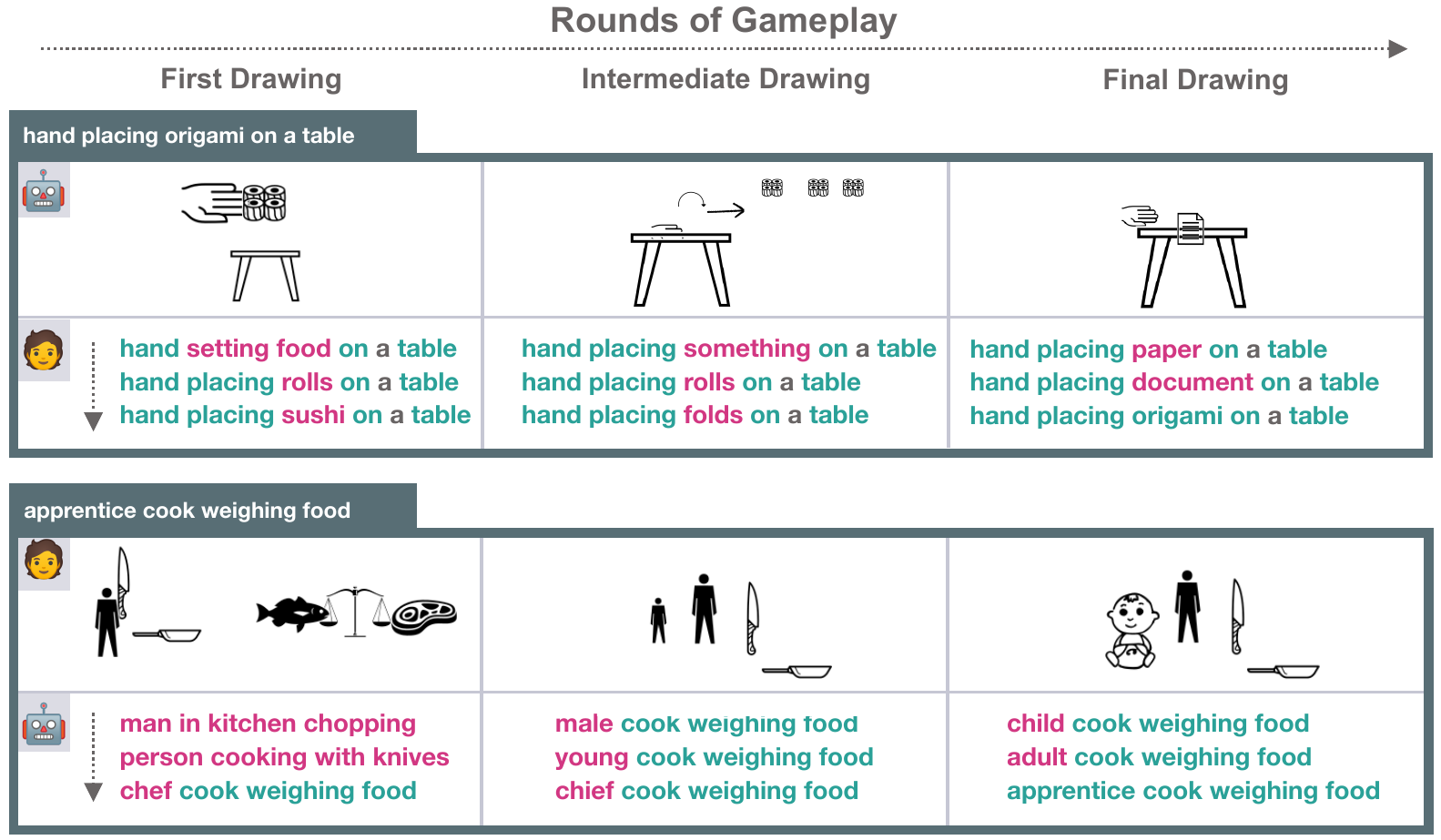}
    \caption{
    Examples of gameplay between human players and our models. Snapshots show the progression (left to right) of two games, with the human player guessing in the top row and drawing in the bottom. Guesses in each round are shown beneath the drawing for that round, and are color-coded (cyan=correctly, magenta=incorrectly guessed word). The first game shows \tdrawer drawing `origami' with a sushi icon (presumably to indicate Japan), a turning icon and finally a paper icon once the human has guessed `folds'. The second game shows \tguesser correctly guessing `apprentice' by interpreting the icons for baby, adult and knife. The words `origami' and `apprentice' \textbf{do not appear} in the training data for either model. See the appendix for more qualitative results.
    }
    \label{fig:example-game}
\end{figure*}

Communicating with humans is a long-standing goal in AI, and has been studied in the context of natural language for decades.  
Many of the key challenges in this task, such as using a shared understanding of the world, commonsense reasoning, and metaphor are, however, not language-specific, but are instead general-purpose tools that humans use when communicating through other modalities as well. 
For example, understanding what \includegraphics[width=.022\textwidth]{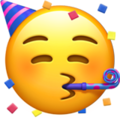} means in a text conversation requires grasping metaphor (it is unlikely to be literally suggesting one should put on a party hat), or understanding a sign with a truck swerving requires common-sense reasoning (the intent is to show slippery conditions, not to suggest drivers ought to begin swerving themselves).
Humans can easily adapt to these different modalities, as well as use visual/symbolic tools (e.g., pointing with a finger, or an arrow in a diagram) that cannot be used in a text-only context. 
To build and test AIs for this skill, we introduce the first task and large-scale dataset for multimodal communication by creating Iconary, a game of drawing and guessing based on Pictionary, along with a dataset of games with human players, proposing automatic and online game playing metrics, and constructing proficient Iconary AIs.

In Iconary, one player (the Drawer) draws an image for a phrase by arranging icons (including the ability to rotate or change the sizes of icons) on a canvas, and a second player (the Guesser) guesses what phrase the drawing represents.
We use icons so we can focus on the high-level semantics of the drawings, and to make the game easier to play online.   
The Guesser then makes a series of attempts to guess the phrase using only the drawing. If the Guesser is unsuccessful, the Drawer can revise the drawing, and the cycle repeats until time runs out or the Guesser is successful.
Figure~\ref{fig:example-game} shows an example of an Iconary game, played between a human player and our AI player.

Iconary combines several key comprehension challenges. First, \textit{non-literal imagery}, since most words in our dataset do not have directly corresponding icons so players will often use visual metaphor (e.g., a school bus and book for `textbook') or reference canonical examples (e.g., lit and unlit light for `turning off') to convey words. 
Second, \textit{visual similarity}, since icons can also be composed to draw objects, such as using concentric circles to draw a dartboard.
Third, \textit{annotations}, because Drawers often use arrows, circles, or crosses to indicate motion or to guide the interpretation of the image.
Fourth, \textit{state tracking}, because players need to remember what drawings/guesses have been already done (e.g., Drawers will often re-draw/augment scenes they could tell confused the Guesser, or use annotations to guide the Guesser's attention towards missed elements).
Fifth, \textit{world knowledge}, since models are tested on words not seen during training.

We present a large dataset for Iconary by having human players play with each other -- a collection of 56k games in train, in-domain (\ind) dev and test sets with 5k games, and out-of-domain (\ood) dev and test sets with 1k and 3k games respectively that contain words not seen during training.

Our proposed models, \tdrawer and \tguesser, leverage world knowledge in the T5~\cite{t5} pre-trained language model and have been carefully adapted to draw and guess words not observed during training. We measure performance using automated metrics, but our main results are shown by having our AIs play games with human players. \tdrawer and \tguesser perform remarkably well on the \ind sets (68.3\% and 96.0\% win rates), but are also able to play impressively with human players on the \ood sets (41.7\% and 62.9\% win rates), demonstrating their ability to extract and integrate world knowledge for unseen game-play words from language models. Figure~\ref{fig:example-game} shows some interesting games played by our models with human partners on the \ood set. 

While our models are capable players, skilled human players outperform them on the \ood sets (a smaller margin of 4.6\% at guessing but a sizeable margin of 21.0\% at drawing). An error analysis shows that most errors occur for unseen words, particularly verbs, compound words, and examples with complex drawings, such as those requiring fine-grained positional information. Our quantitative and qualitative analysis suggests ample room for future research in this new, rich and complex domain. 

\section{The Iconary Game and Dataset}

\begin{figure*}[!t]
\includegraphics[width=\textwidth]{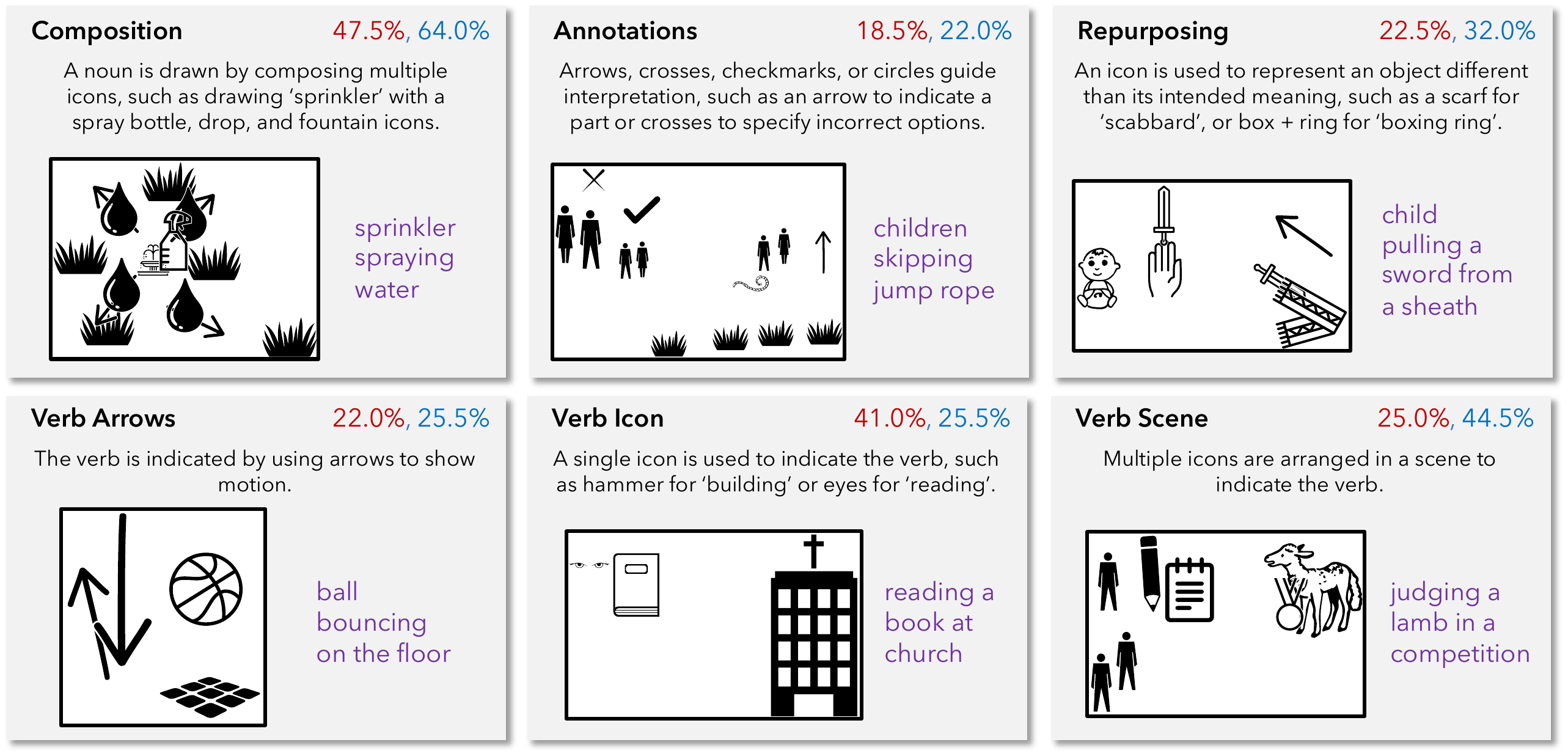}
\caption{Examples of different drawing strategies found in our dataset. The proportion of games that use these methods in a sample from the {\color{red}\ind} and {\color{blue}\ood} dev sets are shown on the top right of each panel.}
\label{fig:drawing-types}
\end{figure*}

\subsection{Playing Iconary}
Iconary is played using a web user interface (UI). First, the Drawer is shown a short phrase and creates a drawing by selecting icons from a library and arranging them on a canvas.
We include 1,205 icons from the Noun Project\footnote{https://thenounproject.com} that were chosen to cover a variety of common entities that would be difficult to draw using other icons. 
Icons can be resized, rotated, and flipped as desired. Once finished, the Drawer passes the turn to the Guesser.

The Guesser is shown the drawing and the phrase with the non-stop words replaced by blanks, and submits a series of guesses to the UI which indicates which words were correct after each guess to allow incremental progress. If the Guesser gives up, control is passed back to the Drawer who can modify their drawing in response to the guesses made so far. This cycle repeats until the phrase is guessed or a 4-minute timeout is reached. The game UI is provided in the appendix.

\subsection{Phrases} 

\begin{table}
    \centering
    \tablefont
\begin{tabular}{ |c|c|c| } 
 \hline
 magnets & doorway & honking \\ 
 swerving & nun & floss \\ 
 roasting & skidding & beverages \\ 
 dreaming & dormitory & librarian \\
 charcoal & cornfield & piloting \\
 rioter & stationary & winery \\ 
 bookmarks & sampling & fireworks \\ 
 lumber & photocopy & shipping \\ 
 unwrapping & freezer & recycling \\
 motorcylist & tidying & waiter \\
 receptionist & pharmacist & stylus \\
 skewers & enchilada & graduating \\
 diet & guitarist & lunchroom \\
 cufflinks & padlocks & soaking \\
 diploma & gunpowder & completing \\
 \hline
\end{tabular}
    \caption{A random sample of 45 \oov words that are present in the \ood dev set, words like `graduating' or `bookmarks' require creativity to draw with icons.}
    \label{tab:oov}
\end{table}

We collect phrases from two sources (see the appendix for more details). First, we have crowdworkers turn image summaries from \imsitu~\cite{imsitu} into short phrases. 
These summaries are derived from FrameNet~\cite{framenet} and consist of an action with the addition of one or more agents (e.g., people, animals), places (e.g., park, office), or artifacts (e.g., computer, car) filling a variety of verb-specific roles. 
We base our phrases on these summaries since they contain words that can be depicted visually, i.e., they avoid abstract words like ``believing" or ``determination" that would be difficult to draw.
We collect 41k phrases with 250 unique verbs, 2k other non-stop words, and an average of 5.4 words. 

Second, we build out-of-domain (\ood) test phrases that have out-of-vocabulary (\oov) words. To maintain the vocabulary size of our training data, we build these phrases by having in-house annotators modify phrases in the \ind test set rather than holding out phrases with particular words from the \imsitu\ phrases. 
First, we collect a list of candidate \oov words by gathering unused words from \imsitu\ and a few other sources, and then manually filtering out words that could not plausibly be drawn. The new \oov words are complex and diverse, see Table~\ref{tab:oov} for a random sample. 
Second, annotators were given a test phrase and asked to write a new phrase that used one of the new words, at least one of the non-stop words from the original phrase, and otherwise preserve as much of the original phrase as possible. 
We build 2.8k new \ood phrases with 1.3k new words. Examples of drawings with these words can be found in the appendix.

The \imsitu\ phrases are divided into train, dev and test sets. Additional filtering was done on dev and test to remove ambiguous words, unusual descriptions and grammatical errors (removing about 15\%). The \ood phrases were divided into dev and test sets, see Table~\ref{tab:dataset-stats} for statistics.

\subsection{Collecting Iconary Games}
We gather Iconary games for these phrases by pairing crowdworkers together to play on our UI. Over 900 players played almost 60,000 games (we allowed multiple games to be played for a phrase). 
Workers qualify by winning a game with another player, and we disqualify workers that have very low win rates during data collection.
We also heuristically filter out poor-quality games, such as removing games with no guesses.
Since the OOD games are our main target, we additionally filter out games with players who had played less than 15 practice games, or that included a small number of players who had win rates far lower than the average, to ensure high quality. 

Table~\ref{tab:dataset-stats} shows statistics for our 5 datasets. 
Humans have a high success rate for the \ind sets. 
The \ood phrases prove more challenging, likely because they often use more advanced words that require more skill to draw and guess.

\begin{table}
    \centering
    \tablefont
    \begin{tabular}{lcccc} \toprule
Dataset & Games & Phrases & Win & Off-by-One \\ \midrule 
Train & 56k & 34k & 71.1 & 83.9\\ 
\ind Valid & 5.1k & 3.1k & 75.1 & 87.5\\ 
\ind Test & 4.7k & 2.9k & 76.8 & 88.3\\ 
\ood Valid & 1.0k & 0.8k & 54.4 & 75.8\\ 
\ood Test & 3.0k & 2.3k & 54.1 & 75.5\\ 
\bottomrule
\end{tabular}
    \caption{Dataset statistics. Off-by-one means the Guesser was within one word of the target phrase.}
    \label{tab:dataset-stats}
\end{table}

\subsection{Analysis}
To better understand our dataset, we perform two analyses. First, we manually label occurrences of six non-exclusive drawing strategies in a sample of 200 games from the \ind and \ood dev sets. 
The results are shown in Figure~\ref{fig:drawing-types}.
We observe that most games use complex strategies to represent the phrase; such as composing multiple icons to represent nouns, drawing small scenes for verbs, using annotations, or creatively re-purposing icons. The \ood dataset tends to include less common nouns and verbs, and drawers adapt to this by using more complex strategies for those phrases. 

Second, we study how Drawers revise their drawings when the Guesser is unsuccessful. We label drawing revisions as either \textbf{edit}: re-arranging, removing, or re-sizing icons, or adding arrows or other annotations, \textbf{add}: adding new icons to offer alternative visualizations or to hint at connections the Drawer missed, \textbf{redraw}: deleting and redrawing parts of a scene that confused the guesser. We make these labels exclusive by placing games into the latter-most category that applies across all drawing revisions in a game. 

The results, and statistics for the use of multiple drawings, are shown in Table~\ref{tab:drawing-revision}. We see that Drawers generally use a balanced mix of our identified strategies and that the more challenging \ood games tend to have more drawings.

\begin{table}
    \centering
    \tablefont
\begin{tabular}{lcccccc} \toprule 
\multirow{2}{*}{Split} & \multicolumn{3}{c}{Rounds} & \multicolumn{3}{c}{Drawing Strategy} \\ 
 & >= 2 & >= 3 & >= 4 & Edit & Add & Redraw \\ \midrule
\ind & 33.3 & 9.4 & 1.9 & 31.5 & 45.0 & 23.5 \\
\ood & 65.6 & 23.8 & 4.5 & 25.5 & 38.5 & 36.0 \\  
\bottomrule
\end{tabular}
    \caption{Statistics for multi-drawing games in the \ind and \ood dev sets. The left three numeric columns show the percent of games with different numbers of drawings, the right three show the usage of different re-drawing strategies.}
    \label{tab:drawing-revision}
\end{table}

\section{Models}
We propose \tguesser and \tdrawer to play Iconary. Both models condition on the current game state, meaning the previous drawings, guesses and, for \tdrawer, the game phrase, and then generate either text to guess the phrase (for \tguesser), or a sequence of special tokens that encode a drawing (for \tdrawer). 

Although this involves a visual modality, we propose to use language models for this task because (1) the icon names can be used to understand the drawing and (2) Iconary often 
requires using word knowledge (e.g., mapping person and thumb icons to `hitchhiking' or milk and ice cream icons to `milkshake') that is known to be captured by these models~\cite{roberts-etal-2020-much}.
To do this, we encode the game state as text and apply the T5~\cite{t5} language model by treating the task as a text-to-text conditional generation task. 
Interestingly, we find vision-and-language (V+L) models~\cite{lxmert,uniter} to be less effective, which might be because current V+L models have inferior language-related abilities~\cite{iki2021effect}, or because models trained on photographic images are not well-suited to understand the non-literal imagery found in Iconary.

\subsection{Guesser}
\label{sect:model-guesser}
\begin{figure}[t]
    \centering
    \includegraphics[width=\linewidth]{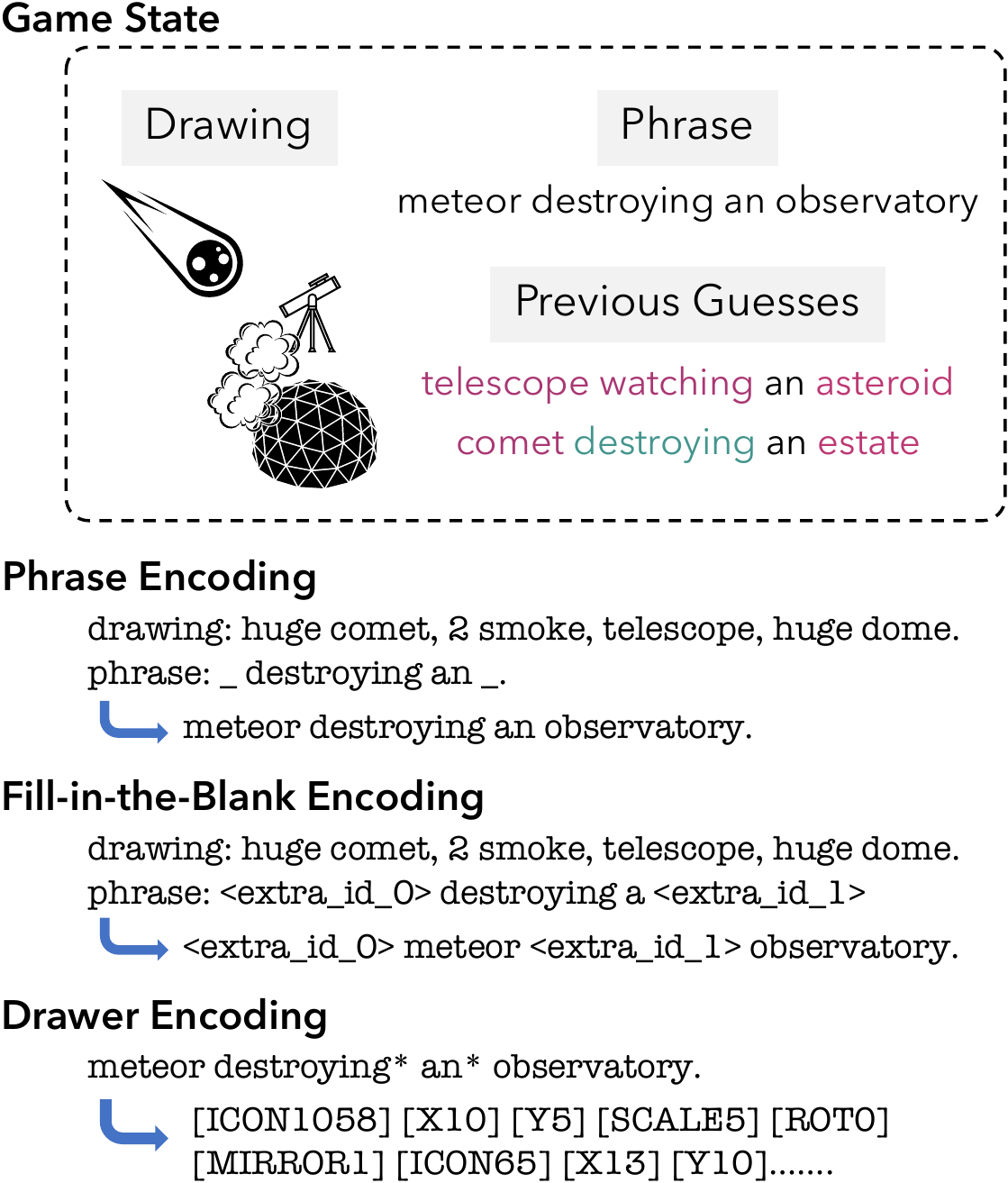}
    \caption{Game state encoding for our models. For each encoding method, the upper text is the input and the lower text is the target output.
    }
    \label{fig:text-encoding}
\end{figure}

To encode the game state for the Guesser, we first construct a text description of the most recent drawing. A description of each icon is built by incorporating the icon name, possibly the prefix `huge', `large', `small' and `tiny' based on the icon's size relative to the other icons, the prefix `rotated' if the icon is rotated, and the prefix `flipped' if the icon is reflected. We handle straight arrows as a special case by encoding them as `[left/right/up/down] arrow' depending on their orientation.
The text description is then a list of these icons sorted from left to right. To keep the result compact for complex scenes, such as a forest drawn with many tree icons, if multiple icons have the same text description we only produce that description once 
and add a number prefix to show the count. 
We use this simplified encoding scheme because preliminary experiments found encoding positional information more precisely, or encoding earlier drawings if they exist, did not improve performance when using T5.

Next, we append the text `phrase:' and, for each word in the target phrase, either an underscore or the correct word if it is known (see Figure~\ref{fig:text-encoding}, top). We experimented with encoding previous incorrect guesses but found it unnecessary as long as models are prevented from repeating those guesses during generation.

The target output is the game phrase. During generation, we constrain models to ensure the output contains the right number of words, includes words that are known to be correct from previous guesses, and exclude words that are known to be incorrect. This is non-trivial for wordpiece models, but we leave details in the appendix.

\subsection{Handling OOV Words}
\label{sect:oov-words}
We observe that naively trained models often generate words seen in the training data even when they do not match the drawing. To combat this, we propose several extensions to \tguesser:

\boldheader{Rare Word Boosting} Based on a method from controlled language generation~\cite{ma-etal-2020-powertransformer,ghosh-etal-2017-affect}, we boost the logit score of wordpieces not seen during training. In particular, we add a fixed value (chosen as a hyperparmeter), to the log-probabilities of those wordpieces and then re-apply the softmax operator to get updated word-piece probabilities during generation.

\boldheader{Fill-in-the-Blank Encoding} Following the T5 pre-training format~\cite{t5}, we encode the phrase using `extra\_id' tokens for sequences of unknown words instead of underscores and train the model to only predict the text that ought to replace those tokens. Figure~\ref{fig:text-encoding} contains an example. We expect this will better enable the model to leverage pre-trained knowledge of unseen words; and this does provide improvements (See Table~\ref{tab:guesser-ablate}).

\boldheader{Early Stopping} We find training for only one epoch beneficial on the \ood sets, possibly because more training causes the model to forget about words learned during pre-training, but are still needed in the \ood test sets, due to catastrophic forgetting~\cite{french1999catastrophic}.

\boldheader{Embed Freezing} The word-piece embeddings are frozen to help ensure the model can effectively use wordpieces that were not in the training data.

\subsection{Drawer}
\label{sect:model-drawer}
The Drawer's input is the game phrase, marked with asterisks to show which words have already been guessed. The output encodes icons with six special tokens, each drawn from a set of new tokens added to T5's vocabulary and initialized with random embeddings, one indicating the icon name, and five indicating the quantized x coordinate, y coordinate, scale, rotation and reflection (quantized with 32, 16, 11, 8 and 2 buckets respectively). The full output is a sequence of such icons (see Figure~\ref{fig:text-encoding}). 
Icons are generated in the order used by the human player (we experimented with other orderings, and found them to be less or equally effective), and we mask the output logits to ensure a valid drawing is produced during generation. 
We propose two additions to help models adapt to this output format:

\boldheader{Special Token Initialization} Icon tokens are initialized by averaging the embeddings of the wordpieces of their names, and quantized tokens are initialized with the embedding of numbers (the first x-coordinate special token is initialized with the embedding for `1', the second for `2', etc.). This gives the model some prior knowledge of what the icons are, and a sense of ordering among the quantized tokens~\cite{wallace-etal-2019-nlp}.  

\boldheader{Constrained Training} The output masking used during generation is applied during training so the model does not need to learn the output format.

\section{Experimental Setup}
In this section, we specify our metrics and baselines. We use T5-3B for \tguesser, but T5-Large for \tdrawer since it generates longer sequences and therefore uses more memory. Other hyperparameters and training details are in the appendix.

\subsection{Human/AI Metrics}
\label{sect:human-ai-metrics}
The best test of Iconary models is playing with human players. 
When playing with human players, AI Guessers make up to 5 guesses a drawing since that is typical for human Guessers. To ensure diverse Drawings from AI Drawers, we sample a drawing from the model's conditional distribution instead of using beam search if beam search yields a drawing with the same icons as a previous drawing (if the sample is still similar to a previous drawing, we use it anyway).
Human players use the same UI and are not told whether they are playing a human or an AI. 

Evaluation is complicated by the fact AIs can make more guesses/drawings than human players since they play faster. To control for this, we measure performance after a fixed number of guesses (for Guessers) and a fixed number of drawings (for Drawers). 
We measure the Win Rate, meaning whether the Guesser correctly guesses the game phrase. We also measure the Soft Win Rate, computed as whether the guesser guesses the exact phrase for phrases of length 2 or less, misses one word or less for phrases of length 3-5, and misses two words or less for phrases with 6 or more words. For \ood games, the game is only considered a soft win if at least one of the unseen words is guessed since that is the focus of our evaluation (denoted as Soft Win$^*$ in tables).

We do not do AI/AI evaluations since we find AI players can often win with drawings that would not be understandable to human players.

\subsection{Automatic Evaluation Metrics}
Gathering human/AI games is challenging since it requires human players with experience playing Iconary. To facilitate automatic evaluation, we propose two metrics for both the Guesser and Drawer that can be computed using human/human games.

\boldheader{Win} Whether the Guesser can win from game states in human/human games.
The Guesser generates five guesses for each drawing in a game where it is allowed to see the previous drawings, previous guesses made for those drawings by the human player, and its own previous guesses. Any word the model generates that does not appear in guesses for previous drawings is considered guessed. The game is won if all words are guessed. Note this is a pessimistic metric because models do not get second chances to guess words after they are identified by the human Guesser, but we expect it to be a reasonable proxy for success in human/AI games.

\boldheader{Soft Win} As above, except we evaluate the Guesser's guessed words on the same soft win metric we use for human/AI games.

\boldheader{Icon F1} Treating drawings as bags of icons, we measure the F1 overlap score between human and computer drawings. We only use the initial drawings for each phrase, and we take the maximum F1 over all human drawings if there are multiple human games for a phrase.

\boldheader{Drawing Perplexity} For models that use the same method of encoding the drawing, we compare the perplexity of each human drawing, averaged over all drawings per game, then averaged over all games in the corpus. 

\subsection{Baselines}
We use the following baselines:

\boldheader{TGuesser-Large/T5Drawer-Base} Identical models but with smaller versions of T5.

\boldheader{BART Guesser/Bart Drawer} Identical models with the BART language model~\cite{bart}. For BART Guesser, we adapt the fill-in-the-blank encoding scheme to generate a copy of the input with the mask tokens replaced, instead of only generating the masked-out tokens, to match BART's pre-training format.

\boldheader{Transformer Guesser/Transformer Drawer} 
We train a transformer-based model~\cite{vaswani2017attention} on this task that does not use a pre-trained language model. 
This model also encodes the drawings as a sequence of special tokens during both decoding and encoding, in which case we find it important to apply a data-augmentation strategy to help the model learn mappings between icons and words they might be used for.
See the appendix for details.

\boldheader{TGuesser-IND} \tguesser without the \ood adaptations specified in Section~\ref{sect:oov-words}.

\section{Results}

\subsection{Human/AI Results}

\begin{figure*}
    \centering
    \includegraphics[width=0.48\textwidth]{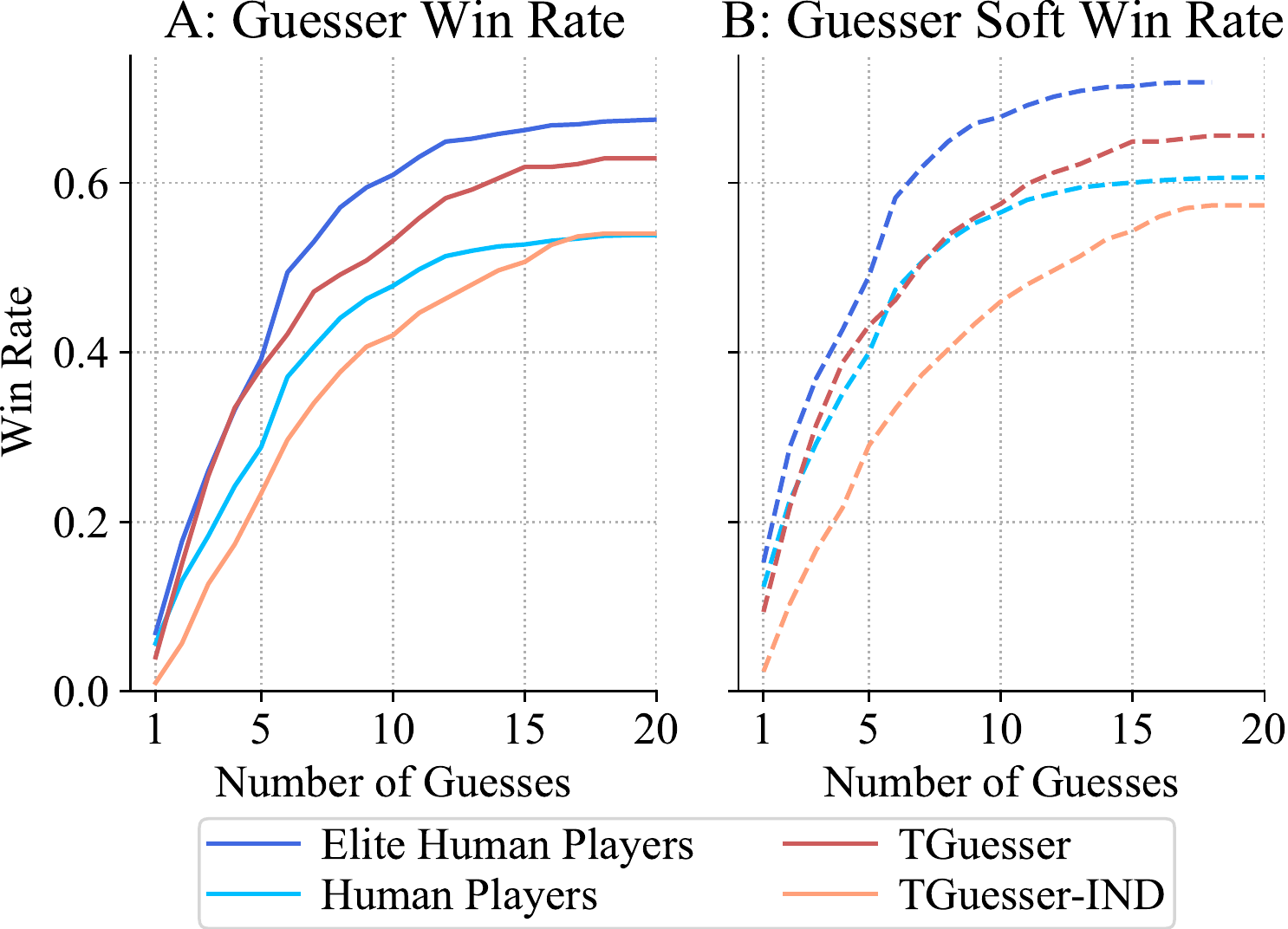}
    \hspace{0.02\textwidth}
    \includegraphics[width=0.48\textwidth]{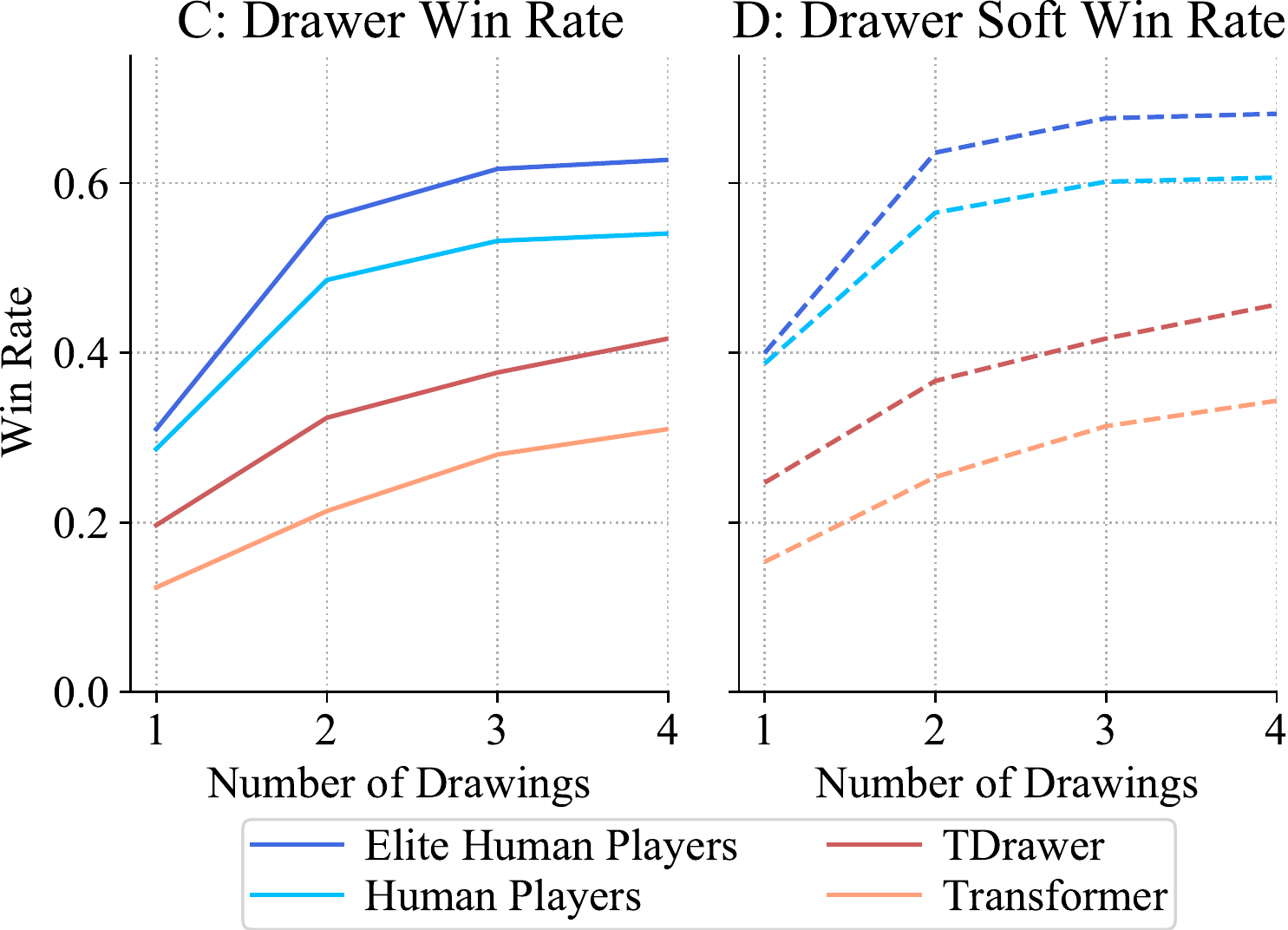}
    \caption{Win rates of our models (\tguesser on the left and \tdrawer on the right) when playing Iconary with human players on phrases from the \ood test set, as more guesses or drawings are used. Graphs with dashed lines show the soft win rate.}
    \label{fig:human-vs-ai}
\end{figure*}

\begin{table}
    \centering
    \tablefont
    \begin{tabular}{lcccc} \toprule 
        \multirow{2}{*}{Model} & \multicolumn{2}{c}{\ind} & \multicolumn{2}{c}{\ood} \\
              & Win & Soft & Win & Soft$^{*}$ \\ \midrule 
\tguesser & 84.25 & 97.62 & \textbf{37.39} & \textbf{44.06}\\ 
\tguesser-IND & 85.91 & 98.55 & 22.67 & 27.24\\ 
\tguesser-Large & 79.34 & 97.09 & 33.30 & 40.61\\ 
BART Guesser & 78.84 & 96.69 & 27.07 & 34.48\\ 
Transformer & 79.89 & 93.64 & 0.00 & 0.00\\ 
\bottomrule
    \end{tabular}
    \caption{Automatic evaluation metrics on the test sets for \tguesser and our baselines.}
    \label{tab:guesser-test}
\end{table}

\setcounter{footnote}{2}
\footnotetext{Due to output format incompatibility, we do not report drawing perplexity for the Transformer baseline.}
\setcounter{footnote}{1}

\begin{table}
    \centering
    \tablefont
    \begin{tabular}{lcccc} \toprule 
        \multirow{2}{*}{Model} & \multicolumn{2}{c}{\ind} & \multicolumn{2}{c}{\ood} \\
              & Icon F1 & Per. & Icon F1 & Per. \\ \midrule 
\tdrawer & 58.04 & 3.84 & \textbf{40.34} & 4.89\\ 
\tdrawer-Base & 58.06 & 3.95 & 39.18 & 5.05\\ 
BART Drawer & 55.07 & 3.67 & 36.64 & \textbf{4.67}\\ 
Transformer & 58.19 & - & 35.78 & -\\ 

\bottomrule
    \end{tabular}
    \caption{Automatic evaluation metrics on the test sets for \tdrawer and our baselines\footnotemark.}
    \label{tab:drawer-test}
\end{table}

Our models and two baselines played 300 games of Iconary with the same crowdworkers used to build our dataset. We evaluate performance on win rate and soft win rate (see Section~\ref{sect:human-ai-metrics}). 
We compare against human/human games, and games with elite human players where either the Guesser (if comparing against an AI Guesser) or Drawer (if comparing against an AI Drawer) is a human player in the top quartile of win rates in human/human games.
We ran experiments on all four models simultaneously, assigning workers to models randomly, and using the same set of 300 phrases randomly selected from the \ood test set for each model.

Results are shown in Figure~\ref{fig:human-vs-ai} (see appendix for tables). We cut off games at 20 guesses for Guessers, and 4 drawings for Drawers, since that is the most human players can typically accomplish in a game (<1\% of human/human games are longer).
At 20 guesses \tguesser has a win rate of 62.9\%, which impressively out-performs the average human player by 9 points, but is still 5 points behind elite human players.
The gap is larger when using the soft win metric, primarily because that metric requires guessing the \oov word, which is unsurprisingly more challenging. There is a large gap between \tguesser and \tguesser-IND, showing our \oov improvements were critical for success.

Drawing is more challenging than guessing. At 4 drawings \tdrawer wins 41.7\% of games, which is significant given the need to draw \oov words. It also outperforms the Transformer baseline suggesting that using T5 did help for \oov words.
Human players, particularly elite players, perform much better, indicating a sizeable opportunity for future research.

We run the same experiment on 300 \ind test phrases using the same pool of annotators, details are in the appendix. We find our models do much better, \tguesser has a win rate of 96.0\% and \tdrawer has a win rate of 68.3\% at 20 guesses and 4 drawings. Human teams on our \ind test and dev sets get 75.9\% for both drawing and guessing. These numbers are not directly comparable since our human/human games used different annotators, but they still make it clear \tguesser is better than human players, and \tdrawer is more comparable to human players, on the \ind phrases.

\subsection{Error Analysis}
We manually annotate 100 unsuccessful games for both \tdrawer and \tguesser (qualitative examples are in the appendix). 
For \tguesser, we find 35\% of errors were on relatively simple scenes where the model guessed related words, but misses the key association. Other errors occur with scenes that used visual similarity (15\%), relied on fine-grained positional information (13\%), had compound words drawn one part at a time (8\%), and other complex scenes (17\%). Only 3\% of cases did not involve the \oov words, and 8\% were clearly deficient drawings.

We find \tdrawer fails to draw anything for \oov words in 32\% of cases, particularly for verbs, possibly because it has learned some verbs do not need cues beyond the related nouns (e.g., `driving' in `person driving a car'). Half the time it draws something related to the \oov words, but that is not sufficient for it to be identified (e.g., `money' for hiring, but without anything to distinguish it from `buy' or `sell').  Only 12\% of unsuccessful games had non-\oov word drawing errors, and 6\% were reasonable drawings.

\subsection{Automatic Evaluation Metrics Results}
We also evaluate our models with automatic metrics on the test sets. Table~\ref{tab:guesser-test} shows the Guesser results. We find that using T5-3B (compared to T5-Large) is quite important. Also, consistent with our human/AI results the \ood optimizations result in a full 15 point gain in performance. The Transformer baseline falls behind the \ind optimized model, and both models on the soft win metric. Its performance is still reasonable, likely because the large training set provides enough examples of humans drawing for it to memorize common drawing strategies or the \ind words. However, the model is unable to learn to predict \ood words (applying \oov boosting for this model only resulted in incoherent output).

Table~\ref{tab:drawer-test} shows the Drawer results. We find \tdrawer benefits somewhat from using a large language model, and that the Transformer baseline is again effective on \ind data but poor on \ood data. BART Drawer shows better perplexity but significantly worse icon overlap.

\subsection{Ablations}

\begin{table}
    \centering
    \tablefont
    \begin{tabular}{lcccc} \toprule 
        \multirow{2}{*}{Model} & \multicolumn{2}{c}{\ind} & \multicolumn{2}{c}{\ood} \\
              & Win & Soft & Win & Soft$^{*}$ \\ \midrule 
\tguesser-Large & 78.96 & 95.92 & \textbf{32.00} & \textbf{39.28}\\ \hdashline
\tguesser-Base & 70.72 & 93.03 & 26.36 & 34.05\\ 
3 Epochs & 82.05 & 96.61 & 29.85 & 34.97\\ 
No Boost & 83.14 & 96.67 & 26.05 & 29.64\\ 
No Fill-in-the-Blank & 82.17 & 96.89 & 29.64 & 34.46\\ 
No Modifiers & 76.77 & 95.06 & 31.49 & 38.67\\ 
Names Only & 74.45 & 94.48 & 29.74 & 35.90\\ 
\bottomrule
    \end{tabular}
    \caption{Guesser ablations on the dev sets. Ablations use T5-Base instead of T5-Large, train for 3 epochs instead of 1, remove \oov boosting, remove fill-in-the-blank encoding, remove modifiers like large/small/rotated from icon names, or use icons names in a randomized order to encode the drawing.}
    \label{tab:guesser-ablate}
\end{table}

We ablate our design choices in more detail using automatic metrics on the dev sets.
Table~\ref{tab:guesser-ablate} shows the Guesser ablations, we use \tguesser-Large to reduce computational expense. Our improvements are impactful with up to 10 points gained through \oov boosting. Icon modifiers help \ind but not \ood, which suggests the model struggles to make use of modifiers for unseen words, however just treating the drawing as a set of icon names clearly harms performance. Fill-in-the-blank encoding is also impactful, suggesting using an encoding scheme similar to the pre-training one is effective for \ood generalization. Unsurprisingly, many of these optimizations reduce \ind performance because they increase the usage \oov words, which never appear in the \ind dev sets. 
Table~\ref{tab:drawer-ablate} shows the Drawer ablations. Our initialization strategy proves to be critical, which suggests it is what allows \tdrawer to leverage the T5 parameter initialization even though it does not output natural language. We also get a modest boost by training with the formatting constraints.

\begin{table}
    \centering
    \tablefont
    \begin{tabular}{lcccc} \toprule 
        \multirow{2}{*}{Model} & \multicolumn{2}{c}{\ind} & \multicolumn{2}{c}{\ood} \\
              & Icon F1 & Per. & Icon F1 & Per. \\ \midrule 
\tdrawer & 57.37 & 3.89 & \textbf{39.96} & \textbf{4.83}\\  \hdashline
\tdrawer-Base & 57.46 & 4.01 & 39.01 & 4.98\\ 
No Icon Init & 47.33 & 4.77 & 31.81 & 5.96\\ 
No Num. Init & 57.04 & 4.09 & 38.58 & 5.05\\ 
No Icon/Num. Init & 44.85 & 4.84 & 28.49 & 5.99\\ 
No Train Const. & 56.06 & 4.12 & 39.17 & 5.20\\ 
\bottomrule
    \end{tabular}
    \caption{Drawer ablations on the dev sets. Ablations use T5-Base instead of T5-Large, remove icon, quantized token, or both initializations, or remove training-time formatting constraints (see  Section~\ref{sect:model-drawer}).
    }
    \label{tab:drawer-ablate}
\end{table}
\section{Related Work}
There is a long history of using games as a testbed for AI. Traditionally these have been adversarial strategy games like Chess~\cite{Silver2018AGR}, Go~\cite{Silver2016MasteringTG}, and many others~\cite{Moravck2017DeepStackEA,Vinyals2017StarCraftIA,mnih2013playing}
A few cooperative games have been studied, like Codenames~\cite{kim2019cooperation} or Hanabi~\cite{walton2019}, that are similar to Iconary in that they require players to communicate in order to achieve a shared goal. 
However, those games severely limit means of communication, whereas Iconary allows a rich variety of communication strategies through the use of drawings, and contains language beyond single words. 
Pictionary-style guessing with freehand drawings has been explored in \citet{sarvadevabhatla2018pictionary,sarvadevabhatla2018game}, although they only consider a single-word single-round setting.

Relating text to visual imagery has also been studied in many forms~\cite{vqa,nlvr}. Generating text that describes visual input, as done in Iconary, has been studied in visual dialog~\cite{visual_dailog}, image captioning~\cite{chen2015microsoft,flickr}, and describing videos ~\cite{aafaq2019video}.
Training models to produce images from text has been studied for captions~\cite{Cho2020XLXMERTPC}, image specifications~\cite{reed2016learning}, and dialogue~\cite{sharma2018chatpainter}.
Unlike in these works, the drawings in Iconary are not photographic and constructed to communicate a phrase. As a result, they can be non-literal and deictic, which makes understanding them a significantly different challenge.  

Using a pre-trained language model to understand mixed language and visual input has been considered by \citet{marasovic-etal-2020-natural}, who use features produced by object detectors or other visual understanding systems as input to GPT-2~\cite{radford2019language} to generate natural language rationales. \citet{scialom-etal-2020-bert} also show BERT~\cite{devlin-etal-2019-bert} can be trained for Visual Question Generation~\cite{vqg}. Similar strategies can be found in many V+L pre-trained models~\cite{lxmert,Lu2019ViLBERTPT,Li2020OscarOA}. We also find combining high-level visual features with a pre-trained language model is an effective way to generate visually relevant text, although again our focus is on drawings rather than photographs.

Figurative text is well studied~\cite{leong-etal-2018-report,veale2016metaphor,shutova-etal-2016-black}, but non-literal imagery has mostly only been explored in the context of parsing charts or diagrams. This includes food webs~\cite{mitra2018knowledge}, science diagrams~\cite{kembhavi2016diagram}, charts~\cite{kafle2018dvqa} or for geometry problems~\cite{seo2014diagram}. 
While this can involve related skills like understanding arrows or using icons to represent concepts, diagrams are usually used to convey technical information and therefore are unlikely to use things like visual metaphor, scenes, or icon compositions to signal words. 

The back-and-forth of Iconary follows a dialogue structure where the Guesser is seeking information from the Drawer. A similar format can be found in dialogue QA datasets~\cite{coqa,quac,aliannejadi2019asking}, and task-oriented dialogue in general similarly requires understanding the intent of a human communicator~\cite{young2013pomdp,chen2017survey}.
Iconary, however, makes this a multimodal process.
\section{Conclusion}
We have presented the game Iconary, a large dataset of human/human games, and our proposed Iconary models. 
This represents the first test for complex multimodal communication between humans and AIs, and is left as an open challenge to the community.

\clearpage

\bibliographystyle{acl_natbib}
\bibliography{main.bib}

\begin{thebibliography}{58}
\expandafter\ifx\csname natexlab\endcsname\relax\def\natexlab#1{#1}\fi

\bibitem[{Aafaq et~al.(2019)Aafaq, Mian, Liu, Gilani, and
  Shah}]{aafaq2019video}
Nayyer Aafaq, Ajmal Mian, Wei Liu, Syed~Zulqarnain Gilani, and Mubarak Shah.
  2019.
\newblock {Video Description: A Survey of Methods, Datasets, and Evaluation
  Metrics}.
\newblock In \emph{ACM Computing Surveys (CSUR)}.

\bibitem[{Aliannejadi et~al.(2019)Aliannejadi, Zamani, Crestani, and
  Croft}]{aliannejadi2019asking}
Mohammad Aliannejadi, Hamed Zamani, Fabio Crestani, and W~Bruce Croft. 2019.
\newblock {Asking Clarifying Questions in Open-Domain Information-Seeking
  Conversations}.
\newblock In \emph{SIGIR}.

\bibitem[{Antol et~al.(2015)Antol, Agrawal, Lu, Mitchell, Batra, Zitnick, and
  Parikh}]{vqa}
Stanislaw Antol, Aishwarya Agrawal, Jiasen Lu, Margaret Mitchell, Dhruv Batra,
  C~Lawrence Zitnick, and Devi Parikh. 2015.
\newblock {Vqa: Visual Question Answering}.
\newblock In \emph{ICV}.

\bibitem[{Baker et~al.(1998)Baker, Fillmore, and Lowe}]{framenet}
Collin~F. Baker, Charles~J. Fillmore, and John~B. Lowe. 1998.
\newblock \href {https://doi.org/10.3115/980845.980860} {{{The Berkeley
  FrameNet Project}}}.
\newblock In \emph{ACL}.

\bibitem[{Chen et~al.(2017)Chen, Liu, Yin, and Tang}]{chen2017survey}
Hongshen Chen, Xiaorui Liu, Dawei Yin, and Jiliang Tang. 2017.
\newblock {A Survey on Dialogue Systems: Recent Advances and New Frontiers}.
\newblock \emph{Acm Sigkdd Explorations Newsletter}.

\bibitem[{Chen et~al.(2015)Chen, Fang, Lin, Vedantam, Gupta, Doll{\'a}r, and
  Zitnick}]{chen2015microsoft}
Xinlei Chen, Hao Fang, Tsung-Yi Lin, Ramakrishna Vedantam, Saurabh Gupta, Piotr
  Doll{\'a}r, and C~Lawrence Zitnick. 2015.
\newblock \href {https://arxiv.org/abs/1504.00325} {{Microsoft COCO Captions:
  Data Collection and Evaluation Server}}.
\newblock \emph{Computing Research Repository}, arXiv:1504.00325.

\bibitem[{Chen et~al.(2020)Chen, Li, Yu, El~Kholy, Ahmed, Gan, Cheng, and
  Liu}]{uniter}
Yen-Chun Chen, Linjie Li, Licheng Yu, Ahmed El~Kholy, Faisal Ahmed, Zhe Gan,
  Yu~Cheng, and Jingjing Liu. 2020.
\newblock {Uniter: UNiversal Image-TExt Representation Learning}.
\newblock In \emph{ECCV}.

\bibitem[{Cho et~al.(2020)Cho, Lu, Schwenk, Hajishirzi, and
  Kembhavi}]{Cho2020XLXMERTPC}
Jaemin Cho, Jiasen Lu, Dustin Schwenk, Hannaneh Hajishirzi, and Aniruddha
  Kembhavi. 2020.
\newblock \href {https://doi.org/10.18653/v1/2020.emnlp-main.707} {{X-LXMERT:
  Paint, Caption and Answer Questions with Multi-Modal Transformers}}.
\newblock In \emph{EMNLP}.

\bibitem[{Choi et~al.(2018)Choi, He, Iyyer, Yatskar, Yih, Choi, Liang, and
  Zettlemoyer}]{quac}
Eunsol Choi, He~He, Mohit Iyyer, Mark Yatskar, Wen-tau Yih, Yejin Choi, Percy
  Liang, and Luke Zettlemoyer. 2018.
\newblock \href {https://doi.org/10.18653/v1/D18-1241} {{{QuAC: Question
  Answering in Context}}}.
\newblock In \emph{EMNLP}.

\bibitem[{Das et~al.(2017)Das, Kottur, Gupta, Singh, Yadav, Moura, Parikh, and
  Batra}]{visual_dailog}
Abhishek Das, Satwik Kottur, Khushi Gupta, Avi Singh, Deshraj Yadav,
  Jos\'e~M.F. Moura, Devi Parikh, and Dhruv Batra. 2017.
\newblock {V}isual {D}ialog.
\newblock In \emph{CVPR}.

\bibitem[{Devlin et~al.(2019)Devlin, Chang, Lee, and
  Toutanova}]{devlin-etal-2019-bert}
Jacob Devlin, Ming-Wei Chang, Kenton Lee, and Kristina Toutanova. 2019.
\newblock \href {https://doi.org/10.18653/v1/N19-1423} {{{BERT: Pre-training of
  Deep Bidirectional Transformers for Language Understanding}}}.
\newblock In \emph{NAACL}.

\bibitem[{Fellbaum(2010)}]{wordnet}
Christiane Fellbaum. 2010.
\newblock {WordNet}.
\newblock In \emph{Theory and applications of ontology: computer applications}.
  Springer.

\bibitem[{French(1999)}]{french1999catastrophic}
Robert~M French. 1999.
\newblock {Catastrophic Forgetting in Connectionist Networks}.
\newblock \emph{Trends in cognitive sciences}.

\bibitem[{Gardner et~al.(2017)Gardner, Grus, Neumann, Tafjord, Dasigi, Liu,
  Peters, Schmitz, and Zettlemoyer}]{allennlp}
Matt Gardner, Joel Grus, Mark Neumann, Oyvind Tafjord, Pradeep Dasigi,
  Nelson~F. Liu, Matthew Peters, Michael Schmitz, and Luke~S. Zettlemoyer.
  2017.
\newblock \href {https://arxiv.org/abs/1803.07640} {{AllenNLP: A Deep Semantic
  Natural Language Processing Platform}}.
\newblock \emph{Computing Research Repository}, arXiv:1803.07640.

\bibitem[{Ghosh et~al.(2017)Ghosh, Chollet, Laksana, Morency, and
  Scherer}]{ghosh-etal-2017-affect}
Sayan Ghosh, Mathieu Chollet, Eugene Laksana, Louis-Philippe Morency, and
  Stefan Scherer. 2017.
\newblock \href {https://doi.org/10.18653/v1/P17-1059} {{{Affect-LM: A Neural
  Language Model for Customizable Affective Text Generation}}}.
\newblock In \emph{ACL}.

\bibitem[{Grave et~al.(2018)Grave, Bojanowski, Gupta, Joulin, and
  Mikolov}]{fasttext}
Edouard Grave, Piotr Bojanowski, Prakhar Gupta, Armand Joulin, and Tomas
  Mikolov. 2018.
\newblock \href {https://www.aclweb.org/anthology/L18-1550} {Learning word
  vectors for 157 languages}.
\newblock In \emph{LREC}.

\bibitem[{Iki and Aizawa(2021)}]{iki2021effect}
Taichi Iki and Akiko Aizawa. 2021.
\newblock \href {https://arxiv.org/abs/2104.08066} {{Effect of
  Vision-and-Language Extensions on Natural Language Understanding in
  Vision-and-Language Models}}.
\newblock \emph{Computing Research Repository}, arXiv:2104.08066.

\bibitem[{Kafle et~al.(2018)Kafle, Price, Cohen, and Kanan}]{kafle2018dvqa}
Kushal Kafle, Brian Price, Scott Cohen, and Christopher Kanan. 2018.
\newblock {DVQA: Understanding Data Visualizations Via Question Answering}.
\newblock In \emph{CVPR}.

\bibitem[{Kembhavi et~al.(2016)Kembhavi, Salvato, Kolve, Seo, Hajishirzi, and
  Farhadi}]{kembhavi2016diagram}
Aniruddha Kembhavi, Mike Salvato, Eric Kolve, Minjoon Seo, Hannaneh Hajishirzi,
  and Ali Farhadi. 2016.
\newblock {A Diagram is Worth a Dozen Images}.
\newblock In \emph{ECCV}.

\bibitem[{Kim et~al.(2019)Kim, Ruzmaykin, Truong, and
  Summerville}]{kim2019cooperation}
Andrew Kim, Maxim Ruzmaykin, Aaron Truong, and Adam Summerville. 2019.
\newblock Cooperation and codenames: Understanding natural language processing
  via codenames.
\newblock In \emph{Proceedings of the AAAI Conference on Artificial
  Intelligence and Interactive Digital Entertainment}.

\bibitem[{Kingma and Ba(2015)}]{adam}
Diederik~P Kingma and Jimmy Ba. 2015.
\newblock {Adam: A Method for Stochastic Optimization}.
\newblock In \emph{ICLR}.

\bibitem[{Leong et~al.(2018)Leong, Beigman~Klebanov, and
  Shutova}]{leong-etal-2018-report}
Chee Wee~(Ben) Leong, Beata Beigman~Klebanov, and Ekaterina Shutova. 2018.
\newblock \href {https://doi.org/10.18653/v1/W18-0907} {{A Report on the 2018
  VUA Metaphor Detection Shared Task}}.
\newblock In \emph{Proceedings of the Workshop on Figurative Language
  Processing}.

\bibitem[{Lewis et~al.(2020)Lewis, Liu, Goyal, Ghazvininejad, Mohamed, Levy,
  Stoyanov, and Zettlemoyer}]{bart}
Mike Lewis, Yinhan Liu, Naman Goyal, Marjan Ghazvininejad, Abdelrahman Mohamed,
  Omer Levy, Veselin Stoyanov, and Luke Zettlemoyer. 2020.
\newblock \href {https://doi.org/10.18653/v1/2020.acl-main.703} {{BART:
  Denoising Sequence-to-Sequence Pre-training for Natural Language Generation,
  Translation, and Comprehension}}.
\newblock In \emph{ACL}.

\bibitem[{Li et~al.(2020)Li, Yin, Li, Hu, Zhang, Zhang, Wang, Hu, Dong, Wei,
  Choi, and Gao}]{Li2020OscarOA}
Xiujun Li, Xi~Yin, Chunyuan Li, Xiaowei Hu, Pengchuan Zhang, Lei Zhang, Lijuan
  Wang, Houdong Hu, Li~Dong, Furu Wei, Yejin Choi, and Jianfeng Gao. 2020.
\newblock {Oscar: Object-Semantics Aligned Pre-training for Vision-Language
  Tasks}.
\newblock In \emph{ECCV}.

\bibitem[{Lu et~al.(2019)Lu, Batra, Parikh, and Lee}]{Lu2019ViLBERTPT}
Jiasen Lu, Dhruv Batra, Devi Parikh, and Stefan Lee. 2019.
\newblock {ViLBERT: Pretraining Task-Agnostic Visiolinguistic Representations
  for Vision-and-Language Tasks}.
\newblock In \emph{NeurIPS}.

\bibitem[{Ma et~al.(2020)Ma, Sap, Rashkin, and
  Choi}]{ma-etal-2020-powertransformer}
Xinyao Ma, Maarten Sap, Hannah Rashkin, and Yejin Choi. 2020.
\newblock \href {https://doi.org/10.18653/v1/2020.emnlp-main.602}
  {{{PowerTransformer: Unsupervised Controllable Revision for Biased Language
  Correction}}}.
\newblock In \emph{EMNLP}.

\bibitem[{Marasovi{\'c} et~al.(2020)Marasovi{\'c}, Bhagavatula, Park, Le~Bras,
  Smith, and Choi}]{marasovic-etal-2020-natural}
Ana Marasovi{\'c}, Chandra Bhagavatula, Jae~sung Park, Ronan Le~Bras, Noah~A.
  Smith, and Yejin Choi. 2020.
\newblock \href {https://doi.org/10.18653/v1/2020.findings-emnlp.253} {{Natural
  Language Rationales with Full-Stack Visual Reasoning: From Pixels to Semantic
  Frames to Commonsense Graphs}}.
\newblock In \emph{EMNLP}.

\bibitem[{Michel et~al.(2011)Michel, Shen, Aiden, Veres, Gray, Pickett,
  Hoiberg, Clancy, Norvig, Orwant et~al.}]{googlebooks}
Jean-Baptiste Michel, Yuan~Kui Shen, Aviva~Presser Aiden, Adrian Veres,
  Matthew~K Gray, Joseph~P Pickett, Dale Hoiberg, Dan Clancy, Peter Norvig, Jon
  Orwant, et~al. 2011.
\newblock {Quantitative Analysis of Culture Using Millions of Digitized Books}.
\newblock \emph{Science}.

\bibitem[{Mitra et~al.(2018)Mitra, Baral, and Clark}]{mitra2018knowledge}
Arindam Mitra, Chitta Baral, and Peter Clark. 2018.
\newblock {Knowledge Representation and Reasoning in Answering Science
  Questions: A Case Study for Food Web Questions}.
\newblock In \emph{Knowledge Representation and Reasoning}.

\bibitem[{Mnih et~al.(2013)Mnih, Kavukcuoglu, Silver, Graves, Antonoglou,
  Wierstra, and Riedmiller}]{mnih2013playing}
Volodymyr Mnih, Koray Kavukcuoglu, David Silver, Alex Graves, Ioannis
  Antonoglou, Daan Wierstra, and Martin Riedmiller. 2013.
\newblock {Playing Atari With Deep Reinforcement Learning}.
\newblock In \emph{NeurIPS Deep Learning Workshop}.

\bibitem[{Moravc{\'i}k et~al.(2017)Moravc{\'i}k, Schmid, Burch, Lis{\'y},
  Morrill, Bard, Davis, Waugh, Johanson, and Bowling}]{Moravck2017DeepStackEA}
Matej Moravc{\'i}k, Martina Schmid, Neil Burch, Viliam Lis{\'y}, Dustin
  Morrill, Nolan Bard, Trevor Davis, Kevin Waugh, Michael Johanson, and
  Michael~H. Bowling. 2017.
\newblock {DeepStack: Expert-level artificial intelligence in heads-up no-limit
  poker}.
\newblock \emph{Science}, 356.

\bibitem[{Mostafazadeh et~al.(2016)Mostafazadeh, Misra, Devlin, Mitchell, He,
  and Vanderwende}]{vqg}
Nasrin Mostafazadeh, Ishan Misra, Jacob Devlin, Margaret Mitchell, Xiaodong He,
  and Lucy Vanderwende. 2016.
\newblock \href {https://doi.org/10.18653/v1/P16-1170} {Generating natural
  questions about an image}.
\newblock In \emph{ACL}.

\bibitem[{Pennington et~al.(2014)Pennington, Socher, and Manning}]{glove}
Jeffrey Pennington, Richard Socher, and Christopher Manning. 2014.
\newblock \href {https://doi.org/10.3115/v1/D14-1162} {{{GloVe: Global Vectors
  for Word Representation}}}.
\newblock In \emph{EMNLP}.

\bibitem[{Radford et~al.(2019)Radford, Wu, Child, Luan, Amodei, and
  Sutskever}]{radford2019language}
Alec Radford, Jeff Wu, Rewon Child, David Luan, Dario Amodei, and Ilya
  Sutskever. 2019.
\newblock {Language Models are Unsupervised Multitask Learners}.
\newblock Technical report, OpenAI.

\bibitem[{Raffel et~al.(2020)Raffel, Shazeer, Roberts, Lee, Narang, Matena,
  Zhou, Li, and Liu}]{t5}
Colin Raffel, Noam Shazeer, Adam Roberts, Katherine Lee, Sharan Narang, Michael
  Matena, Yanqi Zhou, Wei Li, and Peter~J. Liu. 2020.
\newblock \href {http://jmlr.org/papers/v21/20-074.html} {{Exploring the Limits
  of Transfer Learning with a Unified Text-to-Text Transformer}}.
\newblock \emph{JMLR}.

\bibitem[{Reddy et~al.(2019)Reddy, Chen, and Manning}]{coqa}
Siva Reddy, Danqi Chen, and Christopher~D. Manning. 2019.
\newblock \href {https://doi.org/10.1162/tacl_a_00266} {{{CoQA: A
  Conversational Question Answering Challenge}}}.
\newblock In \emph{ACL}.

\bibitem[{Reed et~al.(2016)Reed, Akata, Mohan, Tenka, Schiele, and
  Lee}]{reed2016learning}
Scott Reed, Zeynep Akata, Santosh Mohan, Samuel Tenka, Bernt Schiele, and
  Honglak Lee. 2016.
\newblock {Learning What and Where to Draw}.
\newblock In \emph{NeurIPS}.

\bibitem[{Roberts et~al.(2020)Roberts, Raffel, and
  Shazeer}]{roberts-etal-2020-much}
Adam Roberts, Colin Raffel, and Noam Shazeer. 2020.
\newblock \href {https://doi.org/10.18653/v1/2020.emnlp-main.437} {{How Much
  Knowledge Can You Pack Into the Parameters of a Language Model?}}
\newblock In \emph{EMNLP}.

\bibitem[{Sarvadevabhatla et~al.(2018{\natexlab{a}})Sarvadevabhatla, Surya,
  Mittal, and Babu}]{sarvadevabhatla2018game}
Ravi~Kiran Sarvadevabhatla, Shiv Surya, Trisha Mittal, and R~Venkatesh Babu.
  2018{\natexlab{a}}.
\newblock {Game of Sketches: Deep Recurrent Models of Pictionary-Style Word
  Guessing}.
\newblock In \emph{AAAI}.

\bibitem[{Sarvadevabhatla et~al.(2018{\natexlab{b}})Sarvadevabhatla, Surya,
  Mittal, and Babu}]{sarvadevabhatla2018pictionary}
Ravi~Kiran Sarvadevabhatla, Shiv Surya, Trisha Mittal, and R~Venkatesh Babu.
  2018{\natexlab{b}}.
\newblock {Pictionary-Style Word Guessing on Hand-Drawn Object Sketches:
  Dataset, Analysis and Deep Network Models}.
\newblock \emph{IEEE transactions on pattern analysis and machine
  intelligence}.

\bibitem[{Scialom et~al.(2020)Scialom, Bordes, Dray, Staiano, and
  Gallinari}]{scialom-etal-2020-bert}
Thomas Scialom, Patrick Bordes, Paul-Alexis Dray, Jacopo Staiano, and Patrick
  Gallinari. 2020.
\newblock \href {https://www.aclweb.org/anthology/2020.inlg-1.39} {{What BERT
  Sees: Cross-Modal Transfer for Visual Question Generation}}.
\newblock In \emph{International Conference on Natural Language Generation}.

\bibitem[{Seo et~al.(2014)Seo, Hajishirzi, Farhadi, and
  Etzioni}]{seo2014diagram}
Min~Joon Seo, Hannaneh Hajishirzi, Ali Farhadi, and Oren Etzioni. 2014.
\newblock {Diagram Understanding in Geometry Questions}.
\newblock In \emph{AAAI}.

\bibitem[{Sharma et~al.(2018)Sharma, Suhubdy, Michalski, Kahou, and
  Bengio}]{sharma2018chatpainter}
Shikhar Sharma, Dendi Suhubdy, Vincent Michalski, Samira~Ebrahimi Kahou, and
  Yoshua Bengio. 2018.
\newblock {ChatPainter: Improving Text to Image Generation Using Dialogue}.
\newblock \emph{ICLR Workshop}.

\bibitem[{Shazeer and Stern(2018)}]{adafactor}
Noam Shazeer and Mitchell Stern. 2018.
\newblock {Adafactor: Adaptive Learning Rates with Sublinear Memory Cost}.
\newblock In \emph{ICML}.

\bibitem[{Shutova et~al.(2016)Shutova, Kiela, and
  Maillard}]{shutova-etal-2016-black}
Ekaterina Shutova, Douwe Kiela, and Jean Maillard. 2016.
\newblock \href {https://doi.org/10.18653/v1/N16-1020} {{Black Holes and White
  Rabbits: Metaphor Identification with Visual Features}}.
\newblock In \emph{NAACL}.

\bibitem[{Silver et~al.(2016)Silver, Huang, Maddison, Guez, Sifre, van~den
  Driessche, Schrittwieser, Antonoglou, Panneershelvam, Lanctot, Dieleman,
  Grewe, Nham, Kalchbrenner, Sutskever, Lillicrap, Leach, Kavukcuoglu, Graepel,
  and Hassabis}]{Silver2016MasteringTG}
David Silver, Aja Huang, Chris~J. Maddison, Arthur Guez, Laurent Sifre, George
  van~den Driessche, Julian Schrittwieser, Ioannis Antonoglou, Vedavyas
  Panneershelvam, Marc Lanctot, Sander Dieleman, Dominik Grewe, John Nham, Nal
  Kalchbrenner, Ilya Sutskever, Timothy~P. Lillicrap, Madeleine Leach, Koray
  Kavukcuoglu, Thore Graepel, and Demis Hassabis. 2016.
\newblock {Mastering the game of Go with deep neural networks and tree search}.
\newblock \emph{Nature}, 529.

\bibitem[{Silver et~al.(2018)Silver, Hubert, Schrittwieser, Antonoglou, Lai,
  Guez, Lanctot, Sifre, Kumaran, Graepel, Lillicrap, Simonyan, and
  Hassabis}]{Silver2018AGR}
David Silver, Thomas Hubert, Julian Schrittwieser, Ioannis Antonoglou, Matthew
  Lai, Arthur Guez, Marc Lanctot, Laurent Sifre, Dharshan Kumaran, Thore
  Graepel, Timothy~P. Lillicrap, Karen Simonyan, and Demis Hassabis. 2018.
\newblock {A general reinforcement learning algorithm that masters chess,
  shogi, and Go through self-play}.
\newblock \emph{Science}, 362.

\bibitem[{Suhr et~al.(2019)Suhr, Zhou, Zhang, Zhang, Bai, and Artzi}]{nlvr}
Alane Suhr, Stephanie Zhou, Ally Zhang, Iris Zhang, Huajun Bai, and Yoav Artzi.
  2019.
\newblock \href {https://doi.org/10.18653/v1/P19-1644} {{A Corpus for Reasoning
  about Natural Language Grounded in Photographs}}.
\newblock In \emph{ACL}.

\bibitem[{Tan and Bansal(2019)}]{lxmert}
Hao Tan and Mohit Bansal. 2019.
\newblock \href {https://doi.org/10.18653/v1/D19-1514} {{{LXMERT: Learning
  Cross-Modality Encoder Representations from Transformers}}}.
\newblock In \emph{EMNLP}.

\bibitem[{Vaswani et~al.(2017)Vaswani, Shazeer, Parmar, Uszkoreit, Jones,
  Gomez, Kaiser, and Polosukhin}]{vaswani2017attention}
Ashish Vaswani, Noam Shazeer, Niki Parmar, Jakob Uszkoreit, Llion Jones,
  Aidan~N Gomez, Lukasz Kaiser, and Illia Polosukhin. 2017.
\newblock {Attention is All You Need}.
\newblock In \emph{NeurIPS}.

\bibitem[{Veale et~al.(2016)Veale, Shutova, and Klebanov}]{veale2016metaphor}
Tony Veale, Ekaterina Shutova, and Beata~Beigman Klebanov. 2016.
\newblock {Metaphor: A Computational Perspective}.
\newblock \emph{Synthesis Lectures on Human Language Technologies}.

\bibitem[{Vinyals et~al.(2017)Vinyals, Ewalds, Bartunov, Georgiev, Vezhnevets,
  Yeo, Makhzani, K{\"u}ttler, Agapiou, Schrittwieser, Quan, Gaffney, Petersen,
  Simonyan, Schaul, van Hasselt, Silver, Lillicrap, Calderone, Keet, Brunasso,
  Lawrence, Ekermo, Repp, and Tsing}]{Vinyals2017StarCraftIA}
Oriol Vinyals, Timo Ewalds, Sergey Bartunov, Petko Georgiev, Alexander~Sasha
  Vezhnevets, Michelle Yeo, Alireza Makhzani, Heinrich K{\"u}ttler, John
  Agapiou, Julian Schrittwieser, John Quan, Stephen Gaffney, Stig Petersen,
  Karen Simonyan, Tom Schaul, Hado van Hasselt, David Silver, Timothy~P.
  Lillicrap, Kevin Calderone, Paul Keet, Anthony Brunasso, David Lawrence,
  Anders Ekermo, Jacob Repp, and Rodney Tsing. 2017.
\newblock \href {https://arxiv.org/abs/1708.04782} {{StarCraft II: A New
  Challenge for Reinforcement Learning}}.
\newblock \emph{ArXiv}, abs/1708.04782.

\bibitem[{Wallace et~al.(2019)Wallace, Wang, Li, Singh, and
  Gardner}]{wallace-etal-2019-nlp}
Eric Wallace, Yizhong Wang, Sujian Li, Sameer Singh, and Matt Gardner. 2019.
\newblock \href {https://doi.org/10.18653/v1/D19-1534} {{{Do NLP Models Know
  Numbers? Probing Numeracy in Embeddings}}}.
\newblock In \emph{EMNLP}.

\bibitem[{Walton-Rivers et~al.(2019)Walton-Rivers, Williams, and
  Bartle}]{walton2019}
Joseph Walton-Rivers, Piers~R Williams, and Richard Bartle. 2019.
\newblock {The 2018 Hanabi Competition}.
\newblock In \emph{2019 IEEE Conference on Games}.

\bibitem[{Yatskar et~al.(2016)Yatskar, Zettlemoyer, and Farhadi}]{imsitu}
Mark Yatskar, Luke Zettlemoyer, and Ali Farhadi. 2016.
\newblock {Situation Recognition: Visual Semantic Role Labeling for Image
  Understanding}.
\newblock In \emph{CVPR}.

\bibitem[{Young et~al.(2014)Young, Lai, Hodosh, and Hockenmaier}]{flickr}
Peter Young, Alice Lai, Micah Hodosh, and Julia Hockenmaier. 2014.
\newblock \href {https://doi.org/10.1162/tacl_a_00166} {{From image
  descriptions to visual denotations: New similarity metrics for semantic
  inference over event descriptions}}.
\newblock In \emph{TACL}.

\bibitem[{Young et~al.(2013)Young, Ga{\v{s}}i{\'c}, Thomson, and
  Williams}]{young2013pomdp}
Steve Young, Milica Ga{\v{s}}i{\'c}, Blaise Thomson, and Jason~D Williams.
  2013.
\newblock {Pomdp-Based Statistical Spoken Dialog Systems: A Review}.
\newblock \emph{IEEE}.

\bibitem[{Zellers and Choi(2017)}]{zellers-choi-2017-zero}
Rowan Zellers and Yejin Choi. 2017.
\newblock \href {https://doi.org/10.18653/v1/D17-1099} {{{Zero-Shot Activity
  Recognition with Verb Attribute Induction}}}.
\newblock In \emph{EMNLP}.

\end{thebibliography}

\clearpage

\appendix

\twocolumn[\section*{Appendix - \\{\small Iconary: A Pictionary-based Game for Testing Multimodal Communication with Drawings and Text}}
]

The appendix includes the following sections:\\
\begin{itemize}
\itemsep0em 
    \item Sec~\ref{app:qual} - Qualitative Results
    \item Sec~\ref{app:traindata} - Training Data Characteristics
    \item Sec~\ref{app:oov} - Out of Vocabulary Words
    \item Sec~\ref{app:ui} - Iconary UI
    \item Sec~\ref{app:iconary-phrases} - Constructing iconary phrases
    \item Sec~\ref{app:constrain} - Constraining the Guesser Output
    \item Sec~\ref{app:traindetails} - Training Details
    \item Sec~\ref{app:humanainums} - Table of Human/AI Results
    \item Sec~\ref{app:transformer} - Baseline Transformer Models
\end{itemize}

\setcounter{figure}{0}
\setcounter{table}{0}
\setcounter{footnote}{0}
\setcounter{page}{1}

\section{Qualitative Results}
\label{app:qual}

\begin{figure*}[t]
    \centering
    \includegraphics[width=\textwidth]{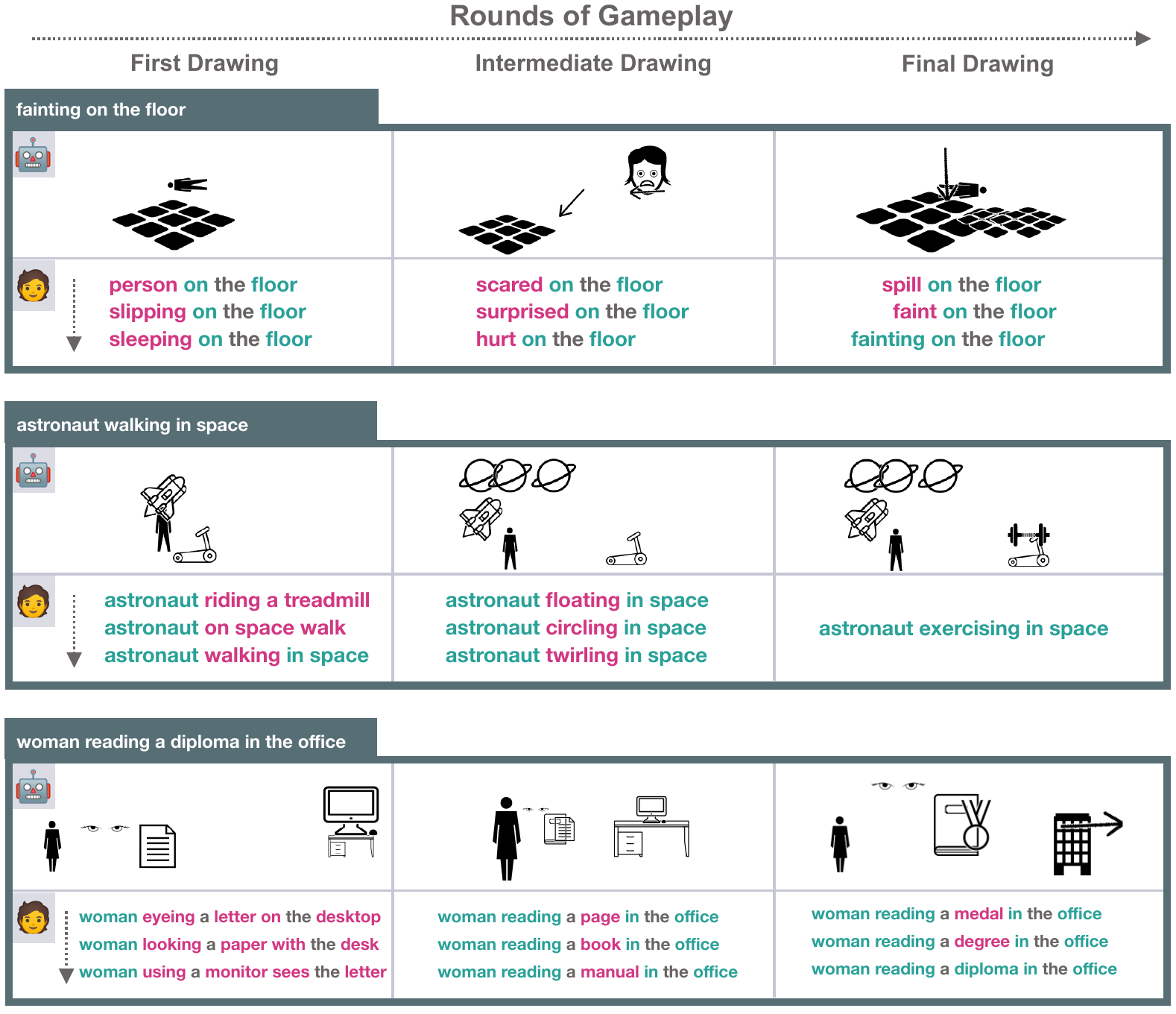}
    \caption{\textbf{\tdrawer qualitative results.} 
    Examples of gameplay between human guessers and \tdrawer. Snapshots show the progression (left to right) of three games. Guesses in each round are shown beneath the drawing for that round and are color-coded (cyan=correctly, magenta=incorrectly guessed word). The first game shows \tdrawer focused on conveying the word 'fainting', a concept not encountered during training. Its first attempt is a literal representing of the phrase, but a subsequent drawing uses a frightened face to convey a possible cause of fainting. The second game shows \tdrawer attempting to draw the unseen word 'astronaut' by using a space shuttle and a ringed planet, which the guesser immediately recognizes.  In the final game  \tdrawer must communicate 'reading a diploma in an office' without having seen the difficult concept of 'diploma' during training. The words `fainting', 'astronaut'  and `diploma' \textbf{do not appear} in the training data for \tdrawer.
    }
    \label{fig:drawer-qualitative} 
\end{figure*}

\begin{figure*}[t]
    \centering
    \includegraphics[width=\textwidth]{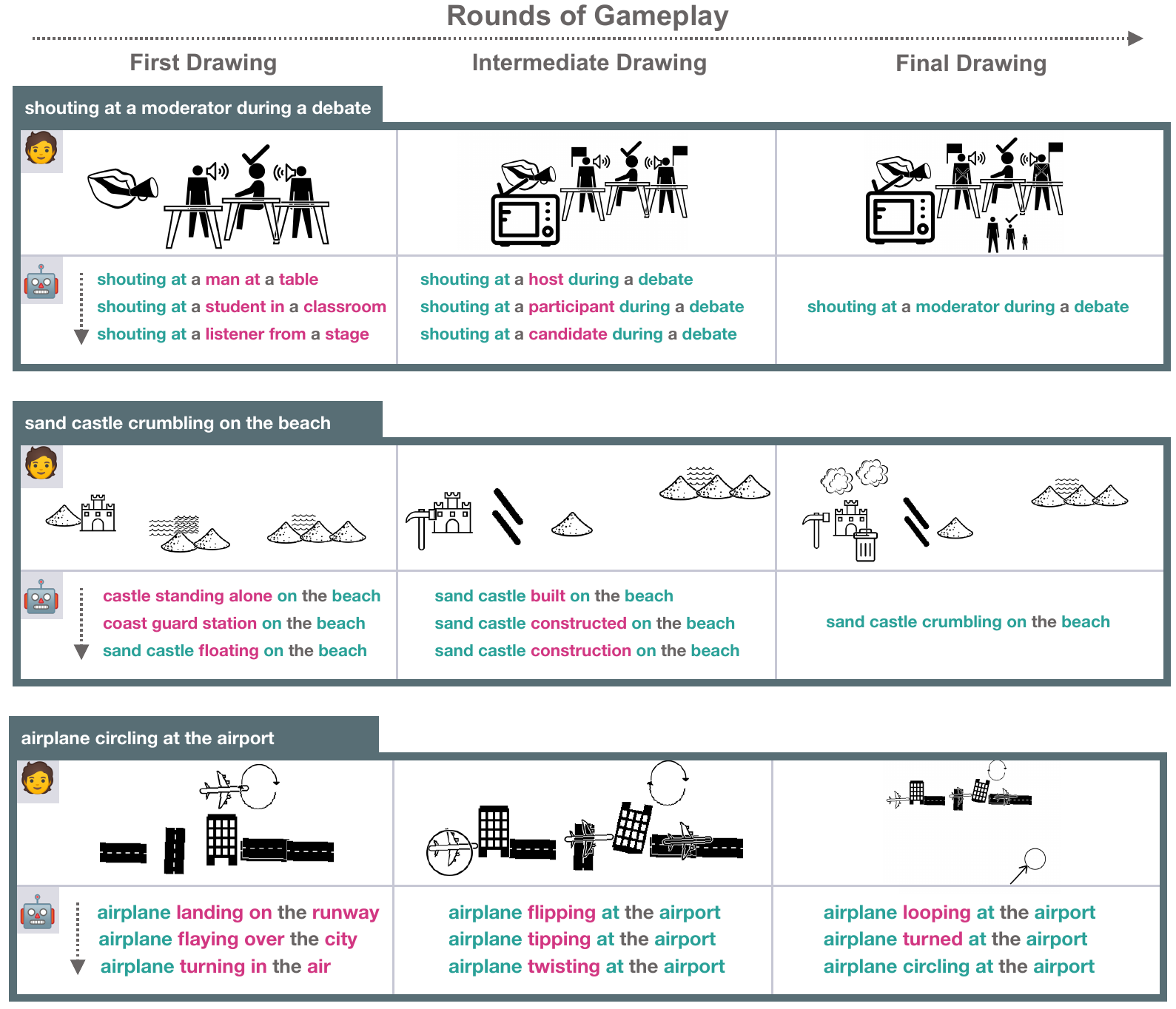}
    \caption{\textbf{\tguesser qualitative results.} 
     Examples of gameplay between \tguesser and human drawers. Snapshots show the progression (left to right) of three games. Guesses in each round are shown beneath the drawing for that round and are color-coded (cyan=correctly, magenta=incorrectly guessed word). In the first game \tguesser quickly gets the action of 'shouting' and the setting of a 'debate', but struggles with the unseen concept of 'moderator' until the human drawer adds a television to their scene. In the second game, the initial drawing is able to convey everything except the unseen verb 'crumbling'. The human drawer is able to use clouds of smoke and a trash can, symbols commonly used for demolition, to get it across. In the last game, the system is unable to guess the unseen verb 'circling' until the human drawer emphasizes the circle icon with an arrow. The words `moderator', 'crumbling'  and `circling' \textbf{do not appear} in the training data for \tguesser.
    }
    \label{fig:guesser-qualitative} 
\end{figure*}

Here we present more qualitative results for human/AI games. Figure~\ref{fig:drawer-qualitative} shows games where the human player guessed the phrase that was drawn by \tdrawer. Figure~\ref{fig:guesser-qualitative} shows games where the human player drew the icon compositions which were then sent to \tguesser to guess.

\section{Training Data Characteristics}
\label{app:traindata}

\begin{figure*}[t]
    \centering
    \includegraphics[width=\textwidth]{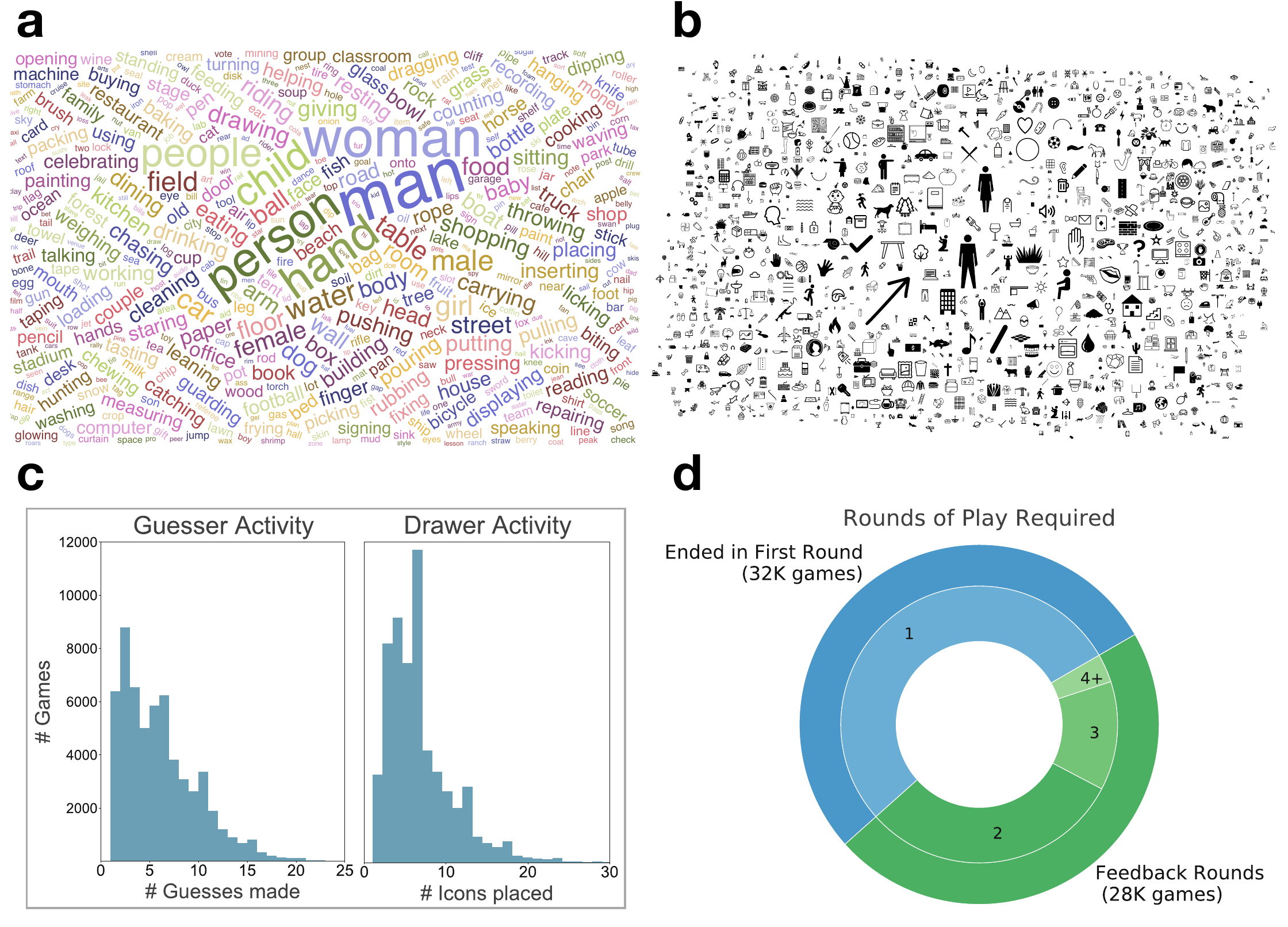}
    \caption{\textbf{Iconary Training Dataset characteristics. a,} 
    Word cloud showing the 500 most frequent words appearing in Iconary phrases, sized by the square root of their relative frequency. \textbf{b,} Cloud of icons available to Iconary players, sized by the square root of their relative frequency. The distributions of words and icons have long tails that contain a rich diversity of concepts. This sparsity forces models to learn concepts and icon usage from a small number of examples. \textbf{c,} Player activity within training set games quantified by the number of guesses and icon placements made by players. A nontrivial number of actions on the part of both players are required for a successful game. \textbf{d,} Breakdown of games by the number of complete rounds of drawing and guessing completed. Nearly half of all games require at least one round of feedback from the drawer, and a significant fraction require multiple rounds.
    }
    \label{fig:training-data-chars} 
\end{figure*}

Figure~\ref{fig:training-data-chars} shows visualizations and statistics for the training dataset used to train \tdrawer and \tguesser. This includes the training word cloud, icon set visualization and activity statistics.

\section{Games with Out of Vocabulary Words}
\label{app:oov}



\begin{figure*}
    \centering
    \includegraphics[width=\textwidth]{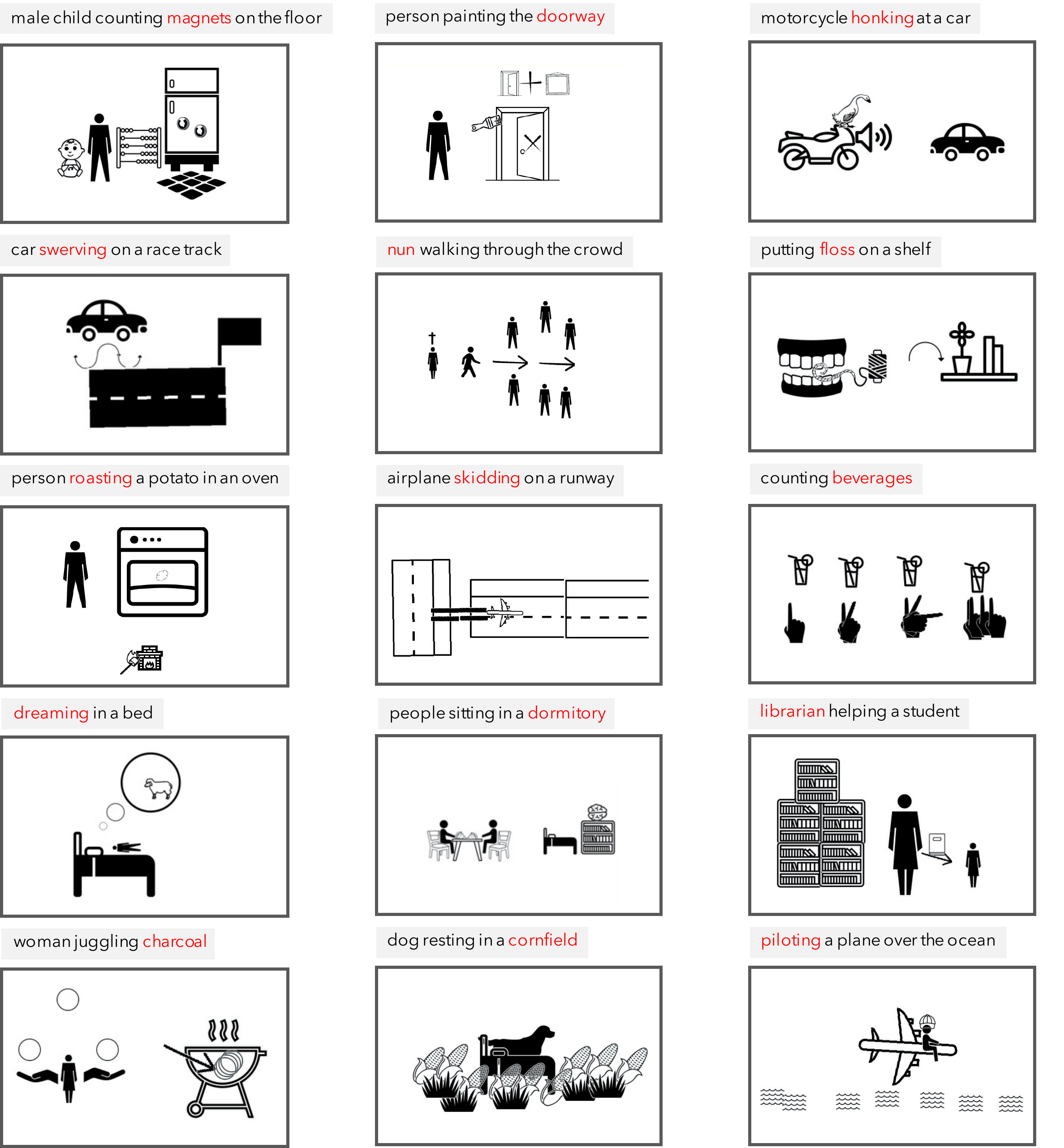}
    \caption{The first drawing for some human-human Iconary games. These phrases belong to the \ood dev set. The {\color{red}word in red} represents the \oov word, not observed in the training set.}
    \label{fig:ood_qual_humans}
\end{figure*}

Figure~\ref{fig:ood_qual_humans} shows the first drawings within games between human players for phrases in the \ood set that contain an \oov word in Table~\ref{tab:oov}. As seen, the drawings for these phrases are rich and often require a creative usage of icons to refer to the \oov words.

\section{Iconary UI}
\label{app:ui}
\begin{figure*}
    \centering
    \includegraphics[width=\textwidth]{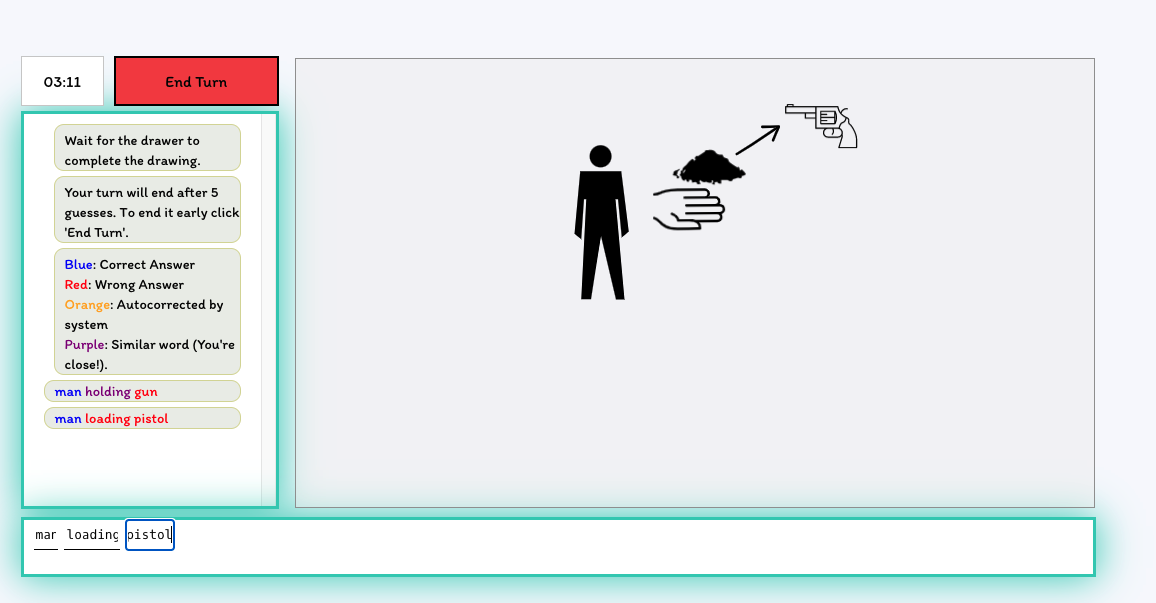}
    \includegraphics[width=\textwidth]{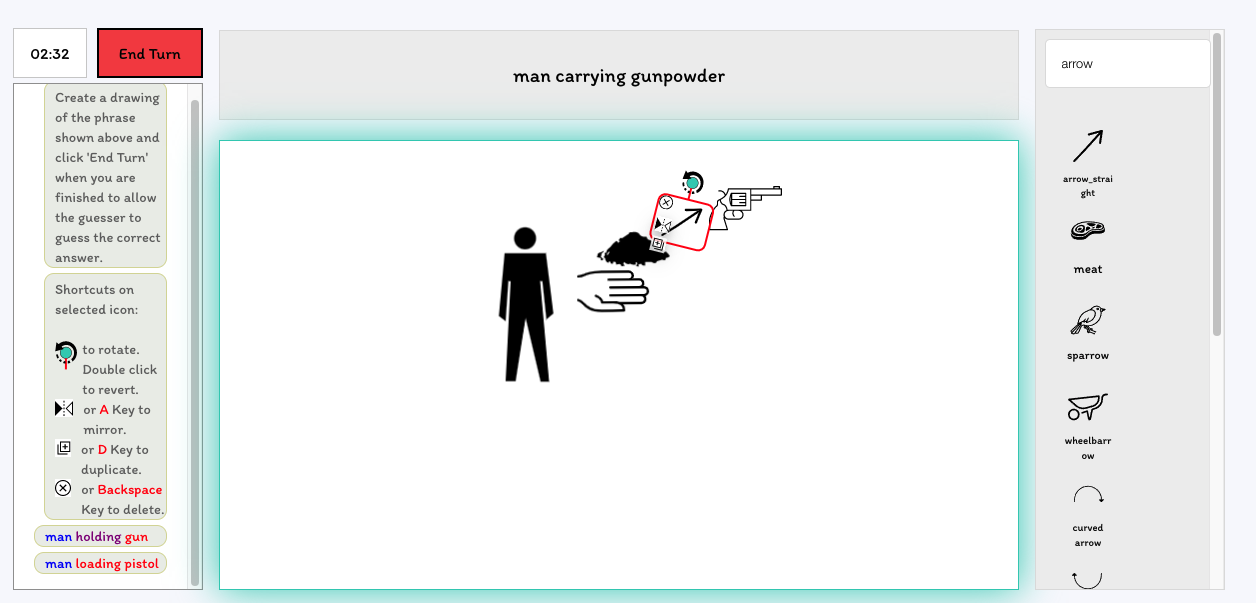}
    \caption{Our UI for playing Iconary. \textbf{Top} shows the Guesser for their first turn of guessing, where they see previous guesses made in the left chatbox, color-coded by whether those guesses were incorrect, correct, or close (judged by word vector similarity). Above that, they see the game time and to the left, the drawing created by the Drawer. At the bottom, the Guesser can enter new guesses by filling in blanks for each word in the phrase. \textbf{Bottom} shows the Drawer on the second turn of drawing. 
    The left panel shows the guesses made by the Guesser and the middle shows the drawing as before. When it is their turn, the Drawer can click on icons to move, resize, rotate, duplicate, delete or reflect them.
    The Drawer can search for icons using text search in the right panel.}
    \label{fig:iconary-ui}
\end{figure*}

Figure~\ref{fig:iconary-ui} shows the UI for playing Iconary.

\section{Constructing Iconary Phrases}
\label{app:iconary-phrases}

In this section, we describe how we build Iconary game phrases in more detail.

\begin{figure*}
    \centering
    \includegraphics[width=\textwidth]{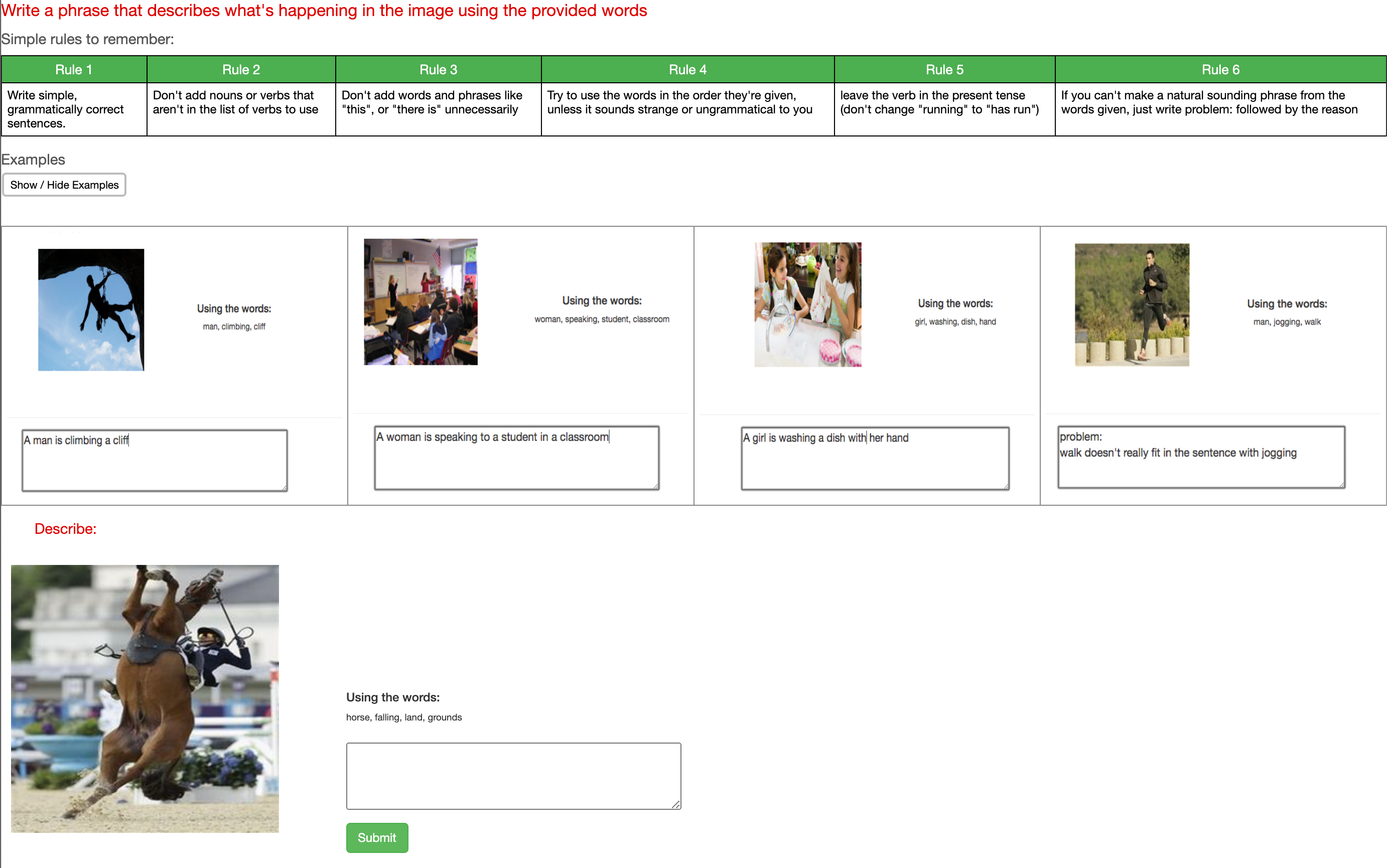}
    \caption{Our UI for authoring Iconary phrases based on the \imsitu\ corpus.}
    \label{fig:imsitu-phrases-ui}
\end{figure*}

\subsection{In-Domain Phrases}
Our primary source of game phrases is derived from the image summaries from the \imsitu\ dataset~\cite{imsitu}.
For each summary, we present crowd workers with the verb, one or more of the associated entities, and ask them to produce a short phrase using those elements. The UI for this task is shown in Figure~\ref{fig:imsitu-phrases-ui}.
We use this process to construct about 41k phrases from 23k frames (a frame can produce multiple phrases depending on the subset of entities used).
Phrases are on average 5.4 words in length and contain 250 unique verbs and 2,000 other non-stop words. 

We hold out 3.5k of these phrases for the \ind test and validation set, ensuring phrases derived from the same \imsitu\ frame are always in the same set. An author of this paper did an additional round of filtering on the test and validation phrases to remove any that contained potentially ambiguous words, described unusual scenes, or contained grammatical errors, leaving 3k phrases for both datasets. The remaining 33k phrases were used for the train set.

\subsection{Collecting Out-of-Domain Phrases}
We also construct a set of out-of-domain (\ood) test phrases that challenge models to play Iconary with out-of-vocabulary (\oov) words. The \imsitu\ data has a limited vocabulary, and building this set by holding out phrases with particular words from the \imsitu\ phrases would further restrict that vocabulary. Instead, we build phrases by having in-house annotators modify phrases in the \ind test set. We consider two kinds of modifications, verb substitutions, and noun substitutions.
\\ \\ \noindent
\textbf{Verb Substitution}: We collect a list of verbs from a variety of sources, including the list of visual verbs from~\citet{zellers-choi-2017-zero}, any verbs in \imsitu\ not already used in the training phrases, and the 1000 most frequent verbs that occur in the Google Books corpus~\cite{googlebooks}. This list was manually filtered to a list of 660 verbs that could plausibly be drawn and do not occur in the original phrase set. Annotators were then given a test phrase and asked to write a new phrase that used one of the new verbs, at least one of the nouns from the original phrase, and otherwise preserve as much of the original phrase as possible. 
\\ \\ \noindent
\textbf{Noun Substitution}: We collect a list of nouns by gathering nouns used in the \imsitu\ corpus that had not yet been used in the training data, and a small number of additional nouns from WordNet~\cite{wordnet} that were not already present, and again manually filter them to ensure they are visually representable. In total, we get 4.6k new nouns.
Annotators were asked to modify a test phrase by re-using the original verb, substituting in one of the new nouns, and otherwise preserving as much of the original phrases as possible. 

In both cases, we make this task easier by building a recommender system that uses the fasttext word vectors~\cite{fasttext} to suggest new noun/verbs that are related to the given phrase.
Altogether, we gather 1.5k new noun phrases and 1.5k new verb phrases that use 1.3k new \oov words.
We reserve a portion of these (0.4k noun and 0.4k verb phrases) for the \ood dev set. 

\section{Constraining the Guesser Output}
\label{app:constrain}
In this section we explain in more detail how we constrain our Guesser wordpiece models to (1) generate the right number of words, (2) always generate known words, and (3) never generate words that are known to be incorrect. The challenge to doing this stems from the fact that these world-level constraints can apply across multiple wordpieces.
We implement 1 and 2 by masking tokens during each generation step, specifically:

\begin{itemize}
    \item If the model is generating a known word, we mask out wordpeices that do not exist in that word and don't start a new word.
    \item If the next word is a known word, we mask out any wordpieces that start new words other than that next known word.
    \item If the word is the last word, we mask out tokens that start a new word, but allow EOS. In other cases, we mask out EOS.
\end{itemize}

This is sufficient to enforce 1 and mostly enforce 2. It is technically possible for the model to only partly generate a known word, or generate some of its wordpeices out-of-order, but models rarely do so in practice because the output would usually be nonsense. 

For 3, we mask out tokens that would start a new word if the word that has just been generated is known to be incorrect. This ensures the model can still generate the wordpieces `run', `er' even if it has already generated `run' as an incorrect guess. This will sometimes mask out all high-probability continuation (e.g., it is unlikely there will be high-probability wordpieces that do not start a new word after generating the word pieces for `runners' if `runners' was an incorrect guess), which can force the model to enter very low-probability generations. To handle this we use a reasonably large number of beams (20), so other beams can be used when this occurs.

Empirically, we find >99.7\% of guess generations from game states in the \ood dev set for \tguesser follow these three constraints.

\section{Training Details}
\label{app:traindetails}
We train our models with Adafactor~\cite{adafactor} with fixed learning rates of 5e-5 for \tguesser and 3e-4 for \tdrawer. 
\tguesser is trained for one epoch as specified in Section~\ref{sect:oov-words} and \tdrawer is trained for two epochs. 

BART Guesser and Drawer are trained with Adam~\cite{adam} with a linearly decreasing learning rates. We train the Guesser for 2 epochs with a learning rate 1e-4, and the Drawer for 3 epochs with a learning rate of 3e-5. Both models linearly warmup the learning from zero for 10\% of the training steps. 

In all cases, we use a batch size of 32. The scale of the \oov boosting was chosen between 0 and 4.0 with increments of 0.5 on the \ood dev set, we use 0.0 for the \tguesser-IND, 3.5 for BART-Guesser, and 2.0 in all other cases. For generation, we use size 20 beam search with the AllenNLP~\cite{allennlp} implementation.

\section{Table of Human/AI Results}
\label{app:humanainums}
In this section, we show Human/AI results in tabular form, as well as the performance of these models when the number of guesses or drawings is unlimited, and our results from the \ind human/AI experiment. 

Table~\ref{tab:human-vs-ai-guesser} shows results for the Guessers, and Table~\ref{tab:human-vs-ai-drawer} shows results for the Drawers from Figure~\ref{fig:human-vs-ai}. The AI players show more improvement if allowed to make more than 20 guesses or 4 drawings than human players, but as stated that is primarily because humans players almost always time-out before reaching that point.

Table~\ref{tab:human-vs-ai-guesser-ind} shows results for the Guessers, and Table~\ref{tab:human-vs-ai-drawer-ind} shows results for the Drawers on our \ind phrases. Note that human performance for these tables is derived from data in the \ind test and dev sets, which used different annotators than the \ood games and our other human/AI experiments, and is therefore not directly comparable. Nevertheless, it is clear \tguesser outperforms humans on these phrases with a win rate close to 100\%, showing that the primary challenge for the Guesser is handling unseen words. \tguesser-IND does slightly better, which is not surprising since it was optimized for \ind performance.

\tdrawer is only slightly behind humans on the \ind phrases, and the Transformer drawer is comparable to humans. The performance improvement is most likely due to the fact models can memorize drawing strategies for different words on the training data, and recompose them for new phrases that reuse those words. It is likely the Transformer Drawer is better able to do this because it was trained on the training data for longer, and the data augmentation strategy in appendix~\ref{appendix:data-augmentation} further guided it towards this approach.

\begin{table*}
    \centering
    \tablefont
\begin{tabular}{lccccccccccc} \toprule
\multirow{2}{*}{Guesser} & \multirow{2}{*}{n} & \multicolumn{2}{c}{5} & \multicolumn{2}{c}{10} & \multicolumn{2}{c}{15} & \multicolumn{2}{c}{20} & \multicolumn{2}{c}{$\infty$} \\
  &   & Win & Soft$^*$ & Win & Soft$^*$ & Win & Soft$^*$ & Win & Soft$^*$ & Win & Soft$^*$ \\ \midrule
Elite Human Players & 888 & 39.19 & 48.99 & 60.92 & 67.79 & 66.22 & 71.40 & 67.45 & 71.85 & 67.57 & 71.85\\ 
Human Players & 3930 & 28.80 & 39.95 & 47.84 & 56.51 & 52.72 & 60.03 & 53.84 & 60.64 & 54.15 & 60.71\\ 
TGuesser & 299 & 38.13 & 43.14 & 53.18 & 57.53 & 61.87 & 64.88 & 62.88 & 65.55 & 66.56 & 68.90\\ 
TGuesser IND & 300 & 23.33 & 29.00 & 42.00 & 46.00 & 50.67 & 54.33 & 54.00 & 57.33 & 60.67 & 63.33\\ \bottomrule
    \end{tabular}
    \caption{Guesser performance from Figure~\ref{fig:human-vs-ai}, left, in tabular form. The top column headers show the number of guesses made, the final column shows performance with an unlimited number of guesses, and the second column shows the number of games we have in each category. One game from \tguesser was removed because a Drawer timed-out without creating a Drawing.}
    \label{tab:human-vs-ai-guesser}
\end{table*}

\begin{table*}
    \centering
    \tablefont
\begin{tabular}{lccccccccccc} \toprule
\multirow{2}{*}{Drawer} & \multirow{2}{*}{n} & \multicolumn{2}{c}{1} & \multicolumn{2}{c}{2} & \multicolumn{2}{c}{3} & \multicolumn{2}{c}{4} & \multicolumn{2}{c}{$\infty$} \\
  &   & Win & Soft$^*$ & Win & Soft$^*$ & Win & Soft$^*$ & Win & Soft$^*$ & Win & Soft$^*$ \\ \midrule
Elite Human Players & 939 & 30.99 & 39.94 & 55.91 & 63.58 & 61.66 & 67.63 & 62.73 & 68.16 & 62.73 & 68.16\\ 
Human Players & 3930 & 28.65 & 38.70 & 48.58 & 56.49 & 53.18 & 60.15 & 54.05 & 60.64 & 54.05 & 60.74\\ 
TDrawer & 300 & 19.67 & 24.67 & 32.33 & 36.67 & 37.67 & 41.67 & 41.67 & 45.67 & 45.00 & 48.33\\ 
Transformer & 300 & 12.33 & 15.33 & 21.33 & 25.33 & 28.00 & 31.33 & 31.00 & 34.33 & 35.00 & 38.33\\   \bottomrule
    \end{tabular}
    \caption{Drawer performance from Figure~\ref{fig:human-vs-ai}, right, in tabular form.}
    \label{tab:human-vs-ai-drawer}
\end{table*}

\begin{table*}
    \centering
    \tablefont
\begin{tabular}{lccccccccccc} \toprule
\multirow{2}{*}{Guesser} & \multirow{2}{*}{n} & \multicolumn{2}{c}{5} & \multicolumn{2}{c}{10} & \multicolumn{2}{c}{15} & \multicolumn{2}{c}{20} & \multicolumn{2}{c}{$\infty$} \\
  &   & Win & Soft & Win & Soft & Win & Soft & Win & Soft & Win & Soft \\ \midrule
Human Players & 9825 & 51.60 & 81.41 & 72.10 & 88.71 & 75.58 & 89.37 & 75.94 & 89.38 & 75.94 & 89.38\\ 
TGuesser & 298 & 83.22 & 98.32 & 93.62 & 98.99 & 95.64 & 98.99 & 95.97 & 98.99 & 95.97 & 98.99\\ 
TGuesser-IND & 300 & 88.00 & 98.67 & 95.33 & 99.33 & 97.67 & 99.67 & 97.67 & 99.67 & 97.67 & 99.67\\   \bottomrule
    \end{tabular}
    \caption{Guesser performance when playing with humans on \ind test phrases.}
    \label{tab:human-vs-ai-guesser-ind}
\end{table*}

\begin{table*}
    \centering
    \tablefont
\begin{tabular}{lccccccccccc} \toprule
\multirow{2}{*}{Drawer} & \multirow{2}{*}{n} & \multicolumn{2}{c}{1} & \multicolumn{2}{c}{2} & \multicolumn{2}{c}{3} & \multicolumn{2}{c}{4} & \multicolumn{2}{c}{$\infty$} \\
  &   & Win & Soft & Win & Soft & Win & Soft$^*$ & Win & Soft & Win & Soft \\ \midrule
Human Players & 9825 & 51.58 & 80.64 & 71.62 & 88.41 & 75.48 & 89.33 & 75.90 & 89.33 & 75.90 & 89.33\\ 
TDrawer & 300 & 39.67 & 75.33 & 59.33 & 88.67 & 66.00 & 88.67 & 68.33 & 88.67 & 69.00 & 88.67\\ 
Transformer & 299 & 45.15 & 74.92 & 62.88 & 87.63 & 68.56 & 89.97 & 71.91 & 91.97 & 73.24 & 91.97\\     \bottomrule
    \end{tabular}
    \caption{Drawer performance when playing with humans on \ind test phrases.}
    \label{tab:human-vs-ai-drawer-ind}
\end{table*}

\section{Transformer Models}
\label{app:transformer}
In the section, we describe our Transformer baselines, which use GloVe~\cite{glove} word embeddings but are otherwise trained from scratch on our training data. Both models use a data augmentation strategy that leverages an icon to word mapping derived from the training data. Both models use 300-dimensional embeddings and 128-dimensional hidden layers, and all hyperparameters were tuned on the \ind dev set.

\subsection{Drawer}
The Transformer Drawer works by encoding the game state and then decoding a drawing in a similar format to \tdrawer. For this model, the last two drawings are converted into the same special tokens used as the output for \tdrawer, which are then embedded with learned embeddings. The game phrase, and the previous guess made by the Guesser if there is one, are also embedded with GloVe word-vectors~\cite{glove}. These elements are concatenated as a sequence and encoded using learned positional embeddings and a 3-layer transformer~\cite{vaswani2017attention}. The decoder is another transformer that cross-attends to the encoded input while generating the output drawing.
The network is optimized with Adam, using a learning rate of $10^{-3}$ for 30 epochs. 

Unlike \tdrawer, the icon ordering for the input and target output is determined by the word-to-icon mapping described in Section~\ref{appendix:data-augmentation}, in particular, icons are ordered in the order of the words they correspond to, and then in the order in which they were drawn. As a result, we are not able to show a comparable perplexity number to \tdrawer in Table~\ref{tab:drawer-test}.

\subsection{Guesser}
The Transformer Guesser is also a conditional generation model. The current drawing, and previous drawing if it exists, are embedded as a sequence using the same quantized format as before. A single transformer then encodes these drawings.

The decoder is a transformer that cross attends to the encoded drawings. We also allow the self-attention layer to attend to future slots in the game phrase, which are filled with the embeddings of the previous guess (or underscores and stopwords if no such guess exists) if those slots occur after the token currently being generated. We use a two-layer multi-layer perceptron with 256 hidden states and ReLU activations to predict the output word.

We again constrain the model to make sure it generates the right number of words, and any known words, during beam search, and select the highest probability beam that did not produce a word known to be incorrect from previous guesses as output. 
This model was trained using Adam~\cite{adam} with a learning rate of $10^{-3}$ for ten epochs, and then with a learning rate for $10^{-5}$ for an additional five epochs.

\subsection{Data Augmentation}
\label{appendix:data-augmentation}
We use data augmentation to boost the performance of both these models (this method did not benefit \tguesser or \tdrawer). First, we derive an icon-to-word mapping from the training data using icon/word co-occurrences by learning icon/word embeddings that are similar for drawings and game phrases found in our data, but dissimilar for drawings paired with random game phrases. Then, for each game, we match icons in drawings for that game to the words in the game phrase that best align with those icons. Finally, we build a pseudo-example by removing some words or constituents from the game phrase and removing the corresponding icons from the drawings. These examples are used as additional training data and are intended to help the models internalize the icon to word co-occurrences that occur in the training data. 

\end{document}